\documentclass[sigconf]{acmart}

\usepackage{subcaption}
\usepackage{algorithm}
\usepackage{algorithmic}
\usepackage{listings}
\usepackage{enumitem}
\usepackage{framed}
\definecolor{codebg}{RGB}{245,245,245}
\definecolor{codegreen}{rgb}{0,0.5,0}
\definecolor{codegray}{rgb}{0.5,0.5,0.5}
\definecolor{codepurple}{rgb}{0.58,0,0.82}
\newcommand{\oa}{\texttt{optimize\_anything}}

\newcommand{\asi}{SI}

\copyrightyear{2026}
\acmYear{2026}
\setcopyright{cc}
\setcctype{by}
\acmConference[CAIS '26]{ACM Conference on AI and Agentic Systems}{May 26--29, 2026}{San Jose, CA, USA}
\acmBooktitle{ACM Conference on AI and Agentic Systems (CAIS '26), May 26--29, 2026, San Jose, CA, USA}
\acmDOI{10.1145/3786335.3813167}
\acmISBN{979-8-4007-2415-2/2026/05}

\acmBadgeR[https://www.acm.org/publications/policies/artifact-review-and-badging-current]{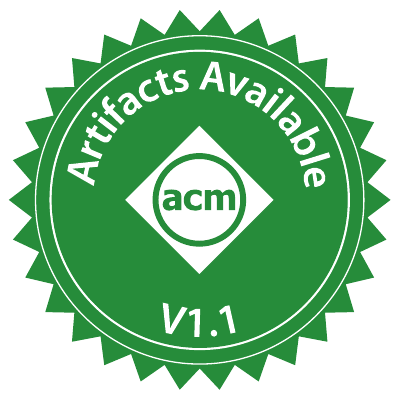}
\acmBadgeR[https://www.acm.org/publications/policies/artifact-review-and-badging-current]{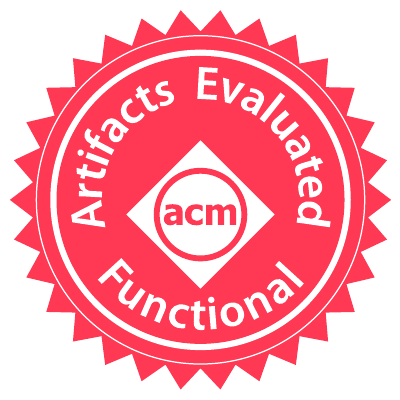}
\acmBadgeR[https://www.acm.org/publications/policies/artifact-review-and-badging-current]{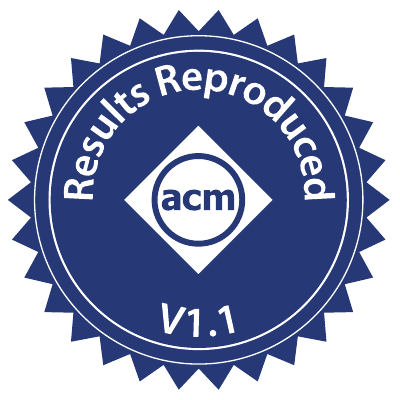}
\makeatletter
\let\@origacmBadgeR\@acmBadgeR
\def\@acmBadgeR{\raisebox{-30pt}[0pt][0pt]{\@origacmBadgeR}}
\makeatother

\definecolor{shadecolor}{gray}{0.95}
\newenvironment{promptbox}[1][]{%
  \par\smallskip
  \def\promptboxtitle{#1}%
  \begin{snugshade}%
  \ifx\promptboxtitle\empty\else
    \noindent\textbf{\small\promptboxtitle}\par\nobreak\smallskip
  \fi
  \small\ttfamily\noindent\ignorespaces
}{%
  \end{snugshade}%
  \par\smallskip
}

\begin{document}

%% ---------- TITLE ----------
\title{\texttt{optimize\_anything}: A Universal API for\\Optimizing any Text Parameter}

\author{Lakshya A Agrawal}
\authornote{Equal contribution.}
\affiliation{\institution{UC Berkeley}\country{USA}}
\email{lakshyaaagrawal@berkeley.edu}

\author{Donghyun Lee}
\authornotemark[1]
\affiliation{\institution{UC Berkeley}\country{USA}}
\email{lukedhlee@berkeley.edu}

\author{Shangyin Tan}
\authornotemark[1]
\affiliation{\institution{UC Berkeley}\country{USA}}
\email{shangyin@berkeley.edu}

\author{Wenjie Ma}
\affiliation{\institution{UC Berkeley}\country{USA}}
\email{windsey@berkeley.edu}

\author{Karim Elmaaroufi}
\affiliation{\institution{UC Berkeley}\country{USA}}
\email{elmaaroufi@berkeley.edu}

\author{Rohit Sandadi}
\affiliation{\institution{UC Berkeley}\country{USA}}
\email{rohitsandadi@berkeley.edu}

\author{Sanjit A. Seshia}
\affiliation{\institution{UC Berkeley}\country{USA}}
\email{sseshia@eecs.berkeley.edu}

\author{Koushik Sen}
\affiliation{\institution{UC Berkeley}\country{USA}}
\email{ksen@cs.berkeley.edu}

\author{Dan Klein}
\affiliation{\institution{UC Berkeley}\country{USA}}
\email{klein@berkeley.edu}

\author{Ion Stoica}
\affiliation{\institution{UC Berkeley}\country{USA}}
\email{istoica@cs.berkeley.edu}

\author{Joseph E.\ Gonzalez}
\affiliation{\institution{UC Berkeley}\country{USA}}
\email{jegonzal@eecs.berkeley.edu}

\author{Omar Khattab}
\affiliation{\institution{MIT}\country{USA}}
\email{okhattab@mit.edu}

\author{Alexandros G. Dimakis}
\affiliation{\institution{UC Berkeley}\country{USA}}
\email{alexdimakis@berkeley.edu}

\author{Matei Zaharia}
\affiliation{\institution{UC Berkeley}\country{USA}}
\email{matei@berkeley.edu}

\renewcommand{\shortauthors}{Agrawal, Lee, Tan, et al.}

\begin{abstract}
Can a single LLM-based optimization system match specialized tools across fundamentally different domains? We show that when optimization problems are formulated as improving a text artifact evaluated by a scoring function, a single AI-based optimization system---supporting single-task search, multi-task search with cross-problem transfer, and generalization to unseen inputs---achieves state-of-the-art results across six diverse tasks. Our system discovers agent architectures that nearly triple Gemini Flash's ARC-AGI accuracy (32.5\% → 89.5\%), finds scheduling algorithms that cut cloud costs by 40\%, generates CUDA kernels where 87\% match or beat PyTorch, and outperforms AlphaEvolve's reported circle packing solution (n=26). Ablations across three domains reveal that actionable side information yields faster convergence and substantially higher final scores than score-only feedback, and that multi-task search outperforms independent optimization given equivalent per-problem budget through cross-task transfer, with benefits scaling with the number of related tasks.
Together, we show for the first time that text optimization with LLM-based search is a general-purpose problem-solving paradigm, unifying tasks traditionally requiring domain-specific algorithms under a single framework. We open-source \oa{} with support for multiple backends as part of the GEPA project at \url{https://github.com/gepa-ai/gepa}.

\end{abstract}




\maketitle

\begin{figure*}[t]
  \centering
  \includegraphics[width=\textwidth]{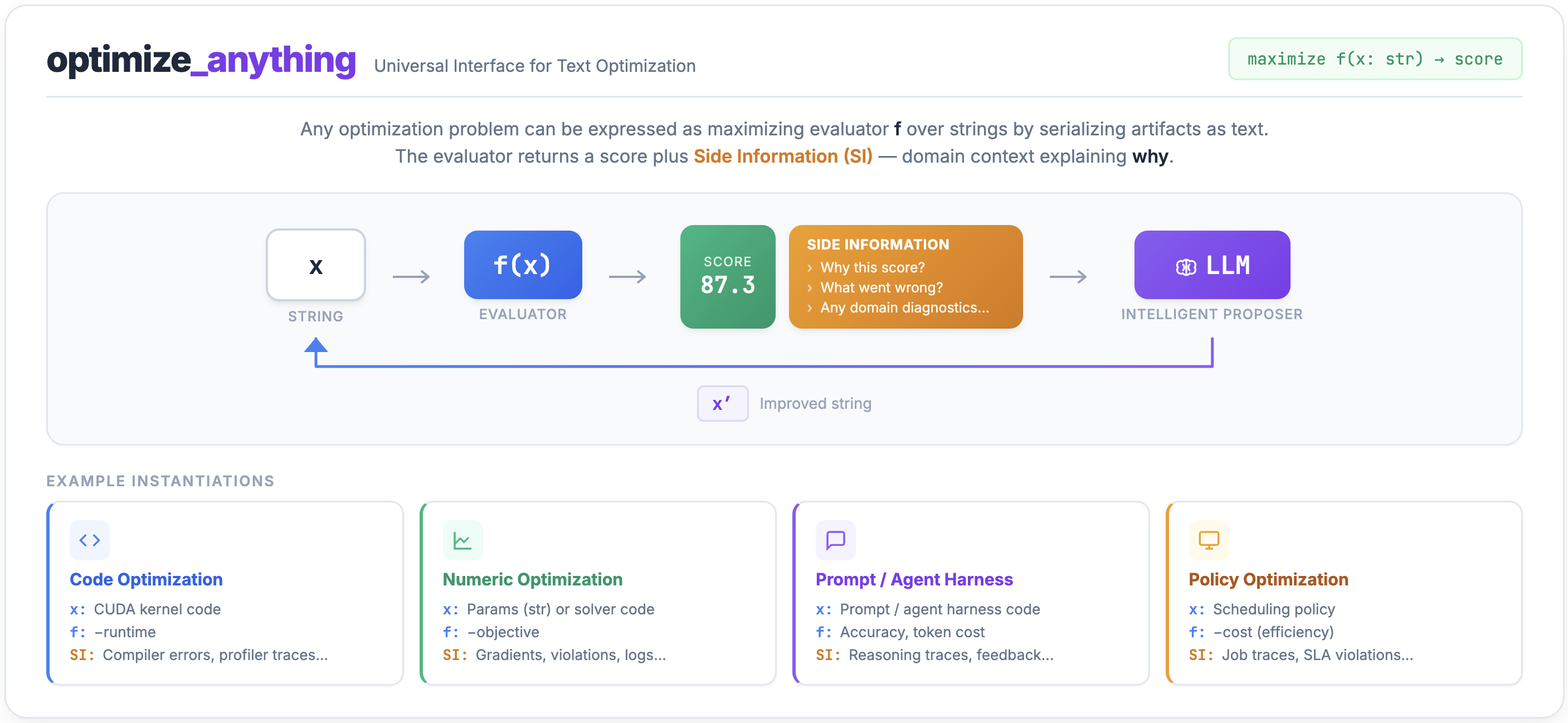}
  \caption{The \oa{} loop: a text artifact $x$ is passed to an evaluator $f(x)$ which returns a score plus diagnostic feedback (\asi{}), which is consumed by an LLM proposer to produce an improved artifact. The same API instantiates across domains: code optimization, prompt tuning, agent architecture search, and policy discovery.}
  \Description{System diagram showing the optimize\_anything loop. A string artifact is evaluated, producing scores and \asi{} feedback, which feeds into an LLM proposer that generates improved candidates. Example instantiations shown for code, prompts, agents, and policies.}
  \label{fig:overview}
\end{figure*}

%% ================================================================
\section{Introduction}\label{sec:intro}

Large language models can serve as effective optimizers when paired with automated evaluation. FunSearch~\cite{romera2024funsearch} evolves Python functions to discover mathematical constructions that surpass known bounds. AlphaEvolve~\cite{alphaevolve2025} extends the idea to broader code optimization, improving a 56-year-old matrix multiplication bound and designing scheduling heuristics for Google's data centers, but it operates exclusively on code artifacts, in single-task mode (one problem at a time).
GEPA~\cite{agrawal2026gepa} achieves state-of-the-art prompt optimization with generalization to unseen inputs, but is limited to prompts; MIPROv2~\cite{opsahl2024miprov2} similarly targets prompt and few-shot selection.
Despite strong results within their artifact types, no existing system has been applied to agent architectures, numeric optimization, or image gen, and no single system has demonstrated effectiveness across fundamentally different domains simultaneously.

We observe that a wide range of problems can be formulated as optimizing a text artifact. Whether the artifact is a CUDA kernel, a cloud scheduling policy, an agent architecture, Scalable Vector Graphics (SVGs), or a system prompt, the structure is the same: serialize the artifact as a string, evaluate it, and let an LLM propose improvements based on diagnostic feedback. This observation suggests a much simpler interface and a uniform algorithm is possible.

We present \oa{} (initially released as ~\citet{agrawal2026oa_blog}), a declarative API that implements this insight. The user provides a seed artifact (or, in seedless mode, just a natural-language objective), an evaluator that returns a score and optional diagnostic feedback, and optionally a dataset. The system handles prompt construction, reflection, candidate selection, and search strategy. This declarative design, inspired by DSPy's~\cite{khattab2023dspy} principle of \emph{programming---not prompting}, means the same API call works whether one is optimizing an LLM prompt, an agent architecture, or an image.

Our contributions are as follows:

\begin{enumerate}[leftmargin=*,itemsep=2pt]
\item \textbf{A single LLM-based Text Optimization system matches or surpasses domain-specific tools across six fundamentally different domains.} We are the first to show that a single system (our proposed \oa{}) can optimize code, prompts, agent architectures, numerical configurations, and images, achieving state-of-the-art results in each. Our system discovers agent architectures that nearly triple ARC-AGI accuracy (32.5\% $\to$ 89.5\%), finds scheduling algorithms that cut cloud costs by 40\%, generates CUDA kernels where 87\% match or beat PyTorch baselines, create custom solver code matching and outperforming Optuna in numerical optimization, and outperforms AlphaEvolve's solution on circle packing. This establishes LLM-based text optimization as a general-purpose problem-solving paradigm, not limited to code or prompts.

\item \textbf{Three optimization modes---single-task, multi-task, and generalization---unified under one interface, including the first multi-task mode.} Existing LLM-evolution systems each support exactly one mode. AlphaEvolve~\cite{alphaevolve2025}, OpenEvolve~\cite{openevolve2025}, and ShinkaEvolve~\cite{shinkaevolve2025} operate in single-task mode: optimizing one code artifact for one problem at a time. GEPA~\cite{agrawal2026gepa} and MIPROv2~\cite{opsahl2024miprov2} operate in generalization mode: optimizing a prompt to perform well on unseen inputs, but only for prompts. No prior system supports \emph{multi-task search}, where solving a batch of related problems together enables cross-transfer of discovered optimization patterns. \oa{} unifies all three modes under one interface: multi-task search on CUDA kernels outperforms independent single-task optimization given equivalent per-problem budget (\S\ref{sec:multi-ablation}), and generalization extends beyond prompts to agent architectures (\S\ref{sec:arc}) and scheduling policies (\S\ref{sec:cloud}). All optimization modes are expressed through the same \oa{} API.

\item \textbf{Side information as a first-class evaluator contract.} Prior frameworks support diagnostic feedback through ad-hoc, framework specific mechanisms. \oa{} elevates it to a uniform API contract: any diagnostic---stack traces, profiler data, rendered images, structured error reports---flows to the proposer through one interface. Ablations across three domains (prompt optimization, circle packing, and CUDA kernels) show that actionable side information yields 4-6$\times$ faster convergence and substantially higher final performance versus score-only feedback (\S\ref{sec:asi-ablation}).
\end{enumerate}

We achieve these results by extending the Pareto-based search of~\citet{agrawal2026gepa} (originally studied only for prompt optimization) to arbitrary text artifacts, adding single-task and multi-task modes. Candidates are selected based on per-example or per-metric Pareto dominance rather than aggregate scores, preserving complementary strengths across iterations. Table~\ref{tab:comparison} provides a detailed comparison.

We evaluate \oa{} across six primary domains spanning all three optimization modes (Table~\ref{tab:summary}), with two additional domains (blackbox mathematical optimization and 3D modeling) in the appendix as preliminary demonstrations. Key results include: (i) evolved agent architectures nearly triple Gemini Flash's ARC-AGI accuracy (32.5\% $\to$ 89.5\%); (ii) discovered cloud scheduling algorithms cut costs by up to 40\%; (iii) 87\% of generated CUDA kernels match or beat PyTorch baselines from KernelBench, with multi-task mode outperforming dedicated single-task optimization; (iv) prompt optimization improves GPT-4.1-mini's AIME-2025 accuracy from 46.67\% to 60.00\%; and (v) our circle packing solution outperforms AlphaEvolve's published one, confirmed by a controlled rerun against OpenEvolve under matched conditions. Ablations across three domains show that actionable side information yields 4-6$\times$ faster convergence and substantially higher final performance versus score-only feedback, and that multi-task search benefits scale with the number of related tasks.

%% ================================================================
\section{Related Work}\label{sec:related}

\paragraph{LLM-based program evolution.}
AlphaEvolve~\cite{alphaevolve2025} pioneered the LLM-evolution paradigm, using Gemini models with island-based MAP-Elites~\cite{mouret2015mapelites} to discover algorithms for Google's infrastructure. OpenEvolve~\cite{openevolve2025} provides an open-source reimplementation with model-agnostic support. ShinkaEvolve~\cite{shinkaevolve2025} extends the paradigm with novelty-based rejection sampling for sample efficiency and adaptive LLM ensemble selection for diversity. FunSearch~\cite{romera2024funsearch} applies evolutionary LLM search to mathematical discovery. EvoPrompting~\cite{chen2024evoprompting} evolves code for neural architecture search. All operate exclusively in single-task mode and expose framework-specific abstractions (island topologies, prompt samplers, evolve-block markers). \oa{} strips the interface to its declarative essence, adds multi-task and generalization modes, and elevates diagnostic feedback to a first-class API concept.

\paragraph{Prompt optimization.}
GEPA~\cite{agrawal2026gepa} combines reflective mutation with a Pareto-based search technique for prompt optimization, outperforming both MIPROv2~\cite{opsahl2024miprov2} and GRPO~\cite{shao2024grpo}. \oa{} supports GEPA's evolutionary search algorithm as one of the optimization backends, extending it beyond prompts to arbitrary text artifacts. Other prompt optimization methods include OPRO~\cite{yang2024opro}, APE~\cite{zhou2023ape}, ProTeGi~\cite{pryzant2023prompt}, and PromptBreeder~\cite{fernando2024promptbreeder}. TextGrad~\cite{yuksekgonul2024textgrad} uses LLM-generated ``gradients'' for text optimization.

\paragraph{LLM self-improvement and reflection.}
Reflexion~\cite{shinn2023reflexion} uses verbal reinforcement for agent self-correction. Self-Refine~\cite{madaan2023selfrefine} applies iterative self-feedback. Evolution through Large Models~\cite{lehman2022evolution} explores LLMs as mutation operators. \oa{}'s \asi{} mechanism generalizes these ideas by making diagnostic feedback a declarative evaluator contract rather than a hardcoded self-critique.

\paragraph{Agent architecture search.}
ADAS~\cite{hu2024automated} and AFlow~\cite{zhang2024aflow} search over agent architectures. \oa{}'s generalization mode subsumes these as special cases: the artifact is the agent code, the evaluator runs it on tasks, and the system evolves both architecture and prompts jointly.

%% ================================================================
\section{The \oa{} API}\label{sec:api}

\subsection{Core Interface}

At its simplest, \oa{} requires a seed artifact and an evaluator. The evaluator takes a candidate string and returns a score (higher is better) alongside an optional Side Information (\asi{}) dictionary containing diagnostic feedback the proposer reads during reflection:

\begin{lstlisting}
import optimize_anything as oa

def evaluate(candidate: str) -> tuple[float, dict]:
    result = execute_code(candidate)
    return result.score, {
        "Error": result.stderr,
        "Output": result.stdout,
        "Runtime": f"{result.time_ms:.1f}ms",
    }

result = oa.optimize_anything(
    seed_candidate="<your artifact>",
    evaluator=evaluate,
)
\end{lstlisting}

\asi{} can include open-ended text, structured data, multiple sub-scores, or images (via \texttt{oa.Image}) for Vision-capable LLMs (VLM).

The full \oa{} signature is:
\begin{lstlisting}
def optimize_anything(
    seed_candidate=None,   # Starting artifact
    evaluator=...,         # fn(candidate) -> Score + SI
    dataset=None,          # Training examples
    valset=None,           # Validation set
    objective=None,        # Natural language goal
    background=None,       # Domain knowledge
    config=None,           # Engine settings
) -> OptimizationResult:
\end{lstlisting}

Specifically, \oa{} doesn't require mutation prompts, task-specific templates, island configurations, or \texttt{EVOLVE-BLOCK} markers (all common in prior frameworks). The user declares the \emph{what} (artifact, evaluator, domain knowledge), and \oa{}, through its optimization backends, handles the \emph{execution}.

\paragraph{Seedless mode.} In domains where providing even a starting artifact is difficult, or where writing even a bad seed requires domain expertise (e.g., 3D modeling), the user can just provide a natural-language \texttt{objective} as an argument in place of the \texttt{seed\_candidate} argument and the LLM bootstraps the first candidate from scratch. Seedless mode makes the system accessible to users who can \emph{specify} what they want but not \emph{implement} it. Appendix~\ref{app:unicorn} demonstrates it on a 3D modeling task.

\subsection{Three Optimization Modes}\label{sec:modes}

Which mode is active depends solely on whether \texttt{dataset} and \texttt{valset} are provided:

\paragraph{Single-Task Search.} No dataset. The candidate \emph{is} the solution; the evaluator scores it directly. This is the mode that AlphaEvolve and OpenEvolve operate in. Example: in circle packing (\S\ref{sec:circle}), the artifact is the packing algorithm and the evaluator returns the packing score plus geometric diagnostics.

\paragraph{Multi-Task Search.} A \texttt{dataset} of related tasks is provided; insights from solving one help solve the others. Example: in CUDA kernel generation (\S\ref{sec:cuda}), each task is a PyTorch operation to accelerate. Multi-task mode discovers optimization patterns that transfer across problems, converging faster and solving more problems than single-task runs (\S\ref{sec:multi-ablation}). No prior LLM-evolution framework supports this mode. Architecturally, the Pareto frontier is shared across tasks for cross-transfer during proposal, but at output time each task independently selects its own best candidate from the frontier. This means multi-task search produces $N$ specialized artifacts (one per task) that have benefited from shared optimization context, patterns discovered while optimizing task $e_i$ are available as parents when proposing for task $e_j$, but each artifact can specialize to its task.

\paragraph{Generalization.} Both \texttt{dataset} and \texttt{valset} are provided; the optimized artifact must perform well on unseen examples. This is the mode that GEPA's prompt optimization~\cite{agrawal2026gepa} operates in;\\ \oa{} generalizes the pattern to any text artifact. Example: in agent architecture discovery (\S\ref{sec:arc}), the artifact is the entire agent, and it must generalize to unseen ARC-AGI puzzles. The key distinction is that multi-task search yields $N$ specialized artifacts while generalization yields one globally generalized artifact.

\begin{table}[t]
\caption{Summary of experimental results across six domains. ``Mode'' indicates which optimization paradigm is used: S = single-task search, M = multi-task search, G = generalization. All results use \oa{} with the indicated proposer LLM.}
\label{tab:summary}
\small
\begin{tabular}{@{}llllr@{}}
\toprule
\textbf{Domain} & \textbf{Mode} & \textbf{Proposer} & \textbf{Key Result} \\
\midrule
Agent Skills (\S\ref{sec:skills}) & G & Claude Opus 4.6 & 100\% pass rate (+20.7pp) \\
Cloud Sched. (\S\ref{sec:cloud}) & G & Gemini 3 Pro & 40.2\% cost savings \\
ARC-AGI (\S\ref{sec:arc}) & G & Gemini 3 Flash & 89.5\% test (+57pp) \\
AIME Prompts (\S\ref{sec:aime}) & G & GPT-5 & 60.0\% test (+13.3pp) \\
CUDA Kernels (\S\ref{sec:cuda}) & M & GPT-5 & 87\% match baseline \\
Circle Pack. (\S\ref{sec:circle}) & S & GPT-5 & Beats AlphaEvolve \\
Math Opt. (\S\ref{app:optuna}) & S & GPT-5 & Wins 7/10 vs Optuna \\
3D Modeling (\S\ref{app:unicorn}) & S & Claude Opus 4.6 & Generates a 3D unicorn \\
\bottomrule
\end{tabular}
\end{table}

%% ================================================================
\section{Method}\label{sec:method}

\oa{} is backend agnostic, and can be used with various optimization algorithms. The default optimization backend in \oa{} currently extends and manages information atop GEPA~\cite{agrawal2026gepa}, an algorithm originally studied primarily in the context of prompt optimization and code search.
The system overview is shown in Figure~\ref{fig:overview}. While \oa{}'s primary contribution is a unified interface, several concrete algorithmic modifications were necessary to generalize from prompts to arbitrary text artifacts: (1)~new frontier types for single-task and multi-task search with distinct selection semantics (GEPA's Pareto-frontier selection relied on evaluation across multiple data points, whereas single-task search admits only one); (2)~a refiner step that catches common LLM generation artifacts (malformed code blocks, import errors, syntax issues) before evaluation, essential for code and agent artifacts where minor formatting errors cause complete evaluation failure; (3)~content-addressed evaluation caching to avoid redundant expensive rollouts; (4)~SI as a first-class typed primitive enabling domain-portable proposer logic and multimodal feedback; and (5)~an adapter layer between various optimization backends and the unified interface.
We describe the two mechanisms that underpin effectiveness and contrast \oa{} with prior frameworks.

\begin{table}[t]
\caption{Comparison of \oa{} with prior LLM-based optimization frameworks across code evolution, prompt optimization, and agent architecture search systems. Only \oa{} supports all three modes and provides diagnostic feedback as a first-class API concept.}
\label{tab:comparison}
\small
\resizebox{\columnwidth}{!}{%
\begin{tabular}{@{}lccccccccc@{}}
\toprule
\textbf{Feature} & \rotatebox{70}{\textbf{AlphaEv.}} & \rotatebox{70}{\textbf{OpenEv.}} & \rotatebox{70}{\textbf{ShinkaEv.}} & \rotatebox{70}{\textbf{GEPA}} & \rotatebox{70}{\textbf{MIPROv2}} & \rotatebox{70}{\textbf{TextGrad}} & \rotatebox{70}{\textbf{ADAS}} & \rotatebox{70}{\textbf{AFlow}} & \rotatebox{70}{\textbf{\oa{}}} \\
\midrule
Code artifacts      & \checkmark & \checkmark & \checkmark &            &            &            &            &            & \checkmark \\
Prompt artifacts    &            &            &            & \checkmark & \checkmark & \checkmark &            &            & \checkmark \\
Agent artifacts     &            &            &            &            &            &            & \checkmark & \checkmark & \checkmark \\
Single-task search  & \checkmark & \checkmark & \checkmark &            &            & \checkmark &            &            & \checkmark \\
Multi-task search   &            &            &            &            &            &            &            &            & \checkmark \\
Generalization      &            &            &            & \checkmark & \checkmark &            & \checkmark & \checkmark & \checkmark \\
\asi{} as API contract &         &            &            &            &            &            &            &            & \checkmark \\
Image feedback      &            &            &            &            &            &            &            &            & \checkmark \\
Declarative interface &          &            &            &            &            &            &            &            & \checkmark \\
No mutation templates &          &            &            & \checkmark & \checkmark & \checkmark & \checkmark & \checkmark & \checkmark \\
Open-source         &            & \checkmark & \checkmark & \checkmark & \checkmark & \checkmark & \checkmark & \checkmark & \checkmark \\
\bottomrule
\end{tabular}%
}
\end{table}

\subsection{Problem Formulation}\label{sec:formulation}

    We formalize the text optimization problem as follows. Let $\mathcal{X}$ denote the space of text artifacts (strings). An evaluator $f:\mathcal{X}\times\mathcal{E} \cup \{\bot\} \to \mathbb{R}\times\mathcal{I}$ maps an artifact $x\in\mathcal{X}$ and an (optional) example $e\in\mathcal{E} \cup \{\bot\} $ to a score $s(x,e)\in\mathbb{R}$ and actionable side information $\iota(x,e)\in\mathcal{I}$, i.e., $f(x,e)=(s(x,e),\iota(x,e))$. The three modes correspond to:

\textbf{Single-task search:} $\mathcal{E}=\emptyset$; maximize $s(x)$ directly. The artifact \emph{is} the solution (e.g., a packing algorithm).

\textbf{Multi-task search:} Given a dataset $\mathcal{D}=\{e_1,\ldots,e_n\}$ of related problems, find an artifact $x\in\mathcal{X}$ (e.g., a kernel-generation prompt) maximizing $\frac{1}{n}\sum_{i=1}^{n} s(x,e_i)$. Cross-transfer arises because the Pareto frontier preserves patterns that work across problems.

\textbf{Generalization:} Given a training set $\mathcal{D}_{\text{train}}$ and a validation set $\mathcal{D}_{\text{val}}=\{e^{\text{val}}_1,\ldots,e^{\text{val}}_k\}$, find an artifact $x\in\mathcal{X}$ maximizing $\frac{1}{k}\sum_{j=1}^{k} s\!\left(x,e^{\text{val}}_j\right)$. Search uses feedback from $\mathcal{D}_{\text{train}}$, while $\mathcal{D}_{\text{val}}$ measures generalization to unseen examples. This generalizes classical machine learning: the artifact may be a prompt, an agent, or a policy.

\subsection{Side Information (\asi{})}\label{sec:asi}

Popularly used numerical optimization methods like gradient descent reduce all diagnostic context to a single scalar. The optimizer knows \emph{that} a candidate failed, but not \emph{why}. For example, one cannot show a Bayesian optimizer a stack trace. LLM-evolution frameworks changed this by feeding execution results into LLM proposers, but when an LLM reads a compiler error, diagnoses a logic bug, and proposes a targeted fix, the process is closer to an engineer iterating on a prototype than to blind evolution.

\oa{} leans into this by making diagnostic feedback a first-class part of the evaluator contract. The evaluator returns both a score and a \texttt{side\_info} dictionary containing any diagnostic the evaluator can produce:

\begin{itemize}[leftmargin=*,itemsep=1pt]
  \item \textbf{Text:} compiler errors, runtime exceptions, profiler summaries, natural-language critiques.
  \item \textbf{Structured data:} per-test-case results, sub-scores for multiple objectives, execution traces.
  \item \textbf{Images:} rendered SVGs, 3D model screenshots, or chart visualizations, enabling VLM proposers to \emph{see} what they are improving.
\end{itemize}

\asi{} is the text-optimization analogue of the gradient. Where gradients tell a numerical optimizer which direction to move, \asi{} can tell the LLM proposer \emph{why} a candidate failed and \emph{how} to fix it. During a dedicated reflection step, the proposer reasons over this signal to diagnose failures and propose targeted improvements.

Prior frameworks expose feedback through framework-specific mechanisms; \asi{} provides a uniform interface that makes it trivial to surface any diagnostic. The key design choice is that \asi{} is \emph{opt-in but zero-friction}: evaluators that return only a score work fine, and existing \texttt{print()} statements can be captured automatically via \texttt{capture\_stdio=True}.

\subsection{Pareto-Based Search}\label{sec:pareto}

Even when optimizing a single objective, evaluating candidates across multiple examples or metrics produces richer signal than a scalar aggregate. The naive approach collapses that signal into one average score and always selects the top candidate. This stalls fast: averaging hides which aspects are strong and which are weak, and the proposer tries to improve everything at once.

\oa{} does two things differently. First, it tracks scores per task (from \texttt{dataset}) or per metric (from sub-scores in \asi{}) individually and maintains a \textbf{Pareto frontier}: any candidate that is the best at \emph{something} survives, even if its average is suboptimal. Second, each reflection step shows the proposer a minibatch of just 2--3 examples instead of all of them, enabling focused, targeted improvements on that subset.

Over iterations, the frontier accumulates complementary strengths. Candidates that excel at different tasks are preserved and their strategies recombined. This mechanism also powers multi-task search: when optimizing across related problems, the frontier preserves candidates that excel on different tasks, and strategies discovered for one problem transfer to others (\S\ref{sec:multi-ablation}).

\paragraph{Candidate selection.} In GEPA~\cite{agrawal2026gepa}, the current default optimization backend, candidates are selected for mutation in proportion to how often they appear on the Pareto front. Let $J$ index the objectives used to form the Pareto scores (e.g., per-example tasks, per-metric scores, or both). Each candidate $\Phi$ induces a score $s_j(\Phi)$ for every $j\in J$. Let $\mathcal{P}$ denote the set of Pareto-nondominated candidates under these objectives. For each objective $j\in J$, let $\mathcal{B}[j]$ be the set of candidates in $\mathcal{P}$ that achieve the best score on $j$. We sample candidates with probability proportional to $|\{j\in J : \Phi \in \mathcal{B}[j]\}|$, focusing exploration on broadly effective solutions.

\paragraph{Reflection and mutation.} Given a selected candidate $\Phi$ and a minibatch $\mathcal{M}$ of examples, the system executes $\Phi$ on $\mathcal{M}$, collects scores and \asi{}, and presents them to the proposer LLM in a structured reflection prompt. The proposer diagnoses failures using the \asi{} and produces an updated artifact $\Phi'$. If $\Phi'$ improves on the minibatch, it is fully evaluated and added to the candidate pool.

%% ================================================================
\section{Experiments}\label{sec:experiments}

We evaluate \oa{} across six domains spanning all three optimization modes. For each, we describe the artifact, evaluator, \asi{} design, and results. We then present ablation studies on multi-task search (\S\ref{sec:multi-ablation}), \asi{} (\S\ref{sec:asi-ablation}), and proposer sensitivity and cost (\S\ref{sec:proposer-sensitivity}), followed by an analysis of the optimization mechanisms (\S\ref{sec:trajectory}).
Optimized solutions are presented in the Appendix \ref{app:discovered_solutions}.

\subsection{Coding Agent Skills (Generalization)}\label{sec:skills}

\textbf{Setup.} Skills are natural-language instructions and best practices for working with a specific codebase (blog post:~\cite{tan2026gskill}). The evaluator runs a coding agent on repository tasks and scores whether it resolves them; the optimized skills must generalize to unseen tasks. We optimize skills for the Bleve search library and evaluate transfer to Claude Code with both Haiku 4.5 and Sonnet 4.5.

\textbf{\asi{} design.} The evaluator returns task descriptions, agent traces (tool calls, code edits, errors), test outcomes, and resolution time.

\begin{figure}[t]
  \centering
  \includegraphics[width=\columnwidth]{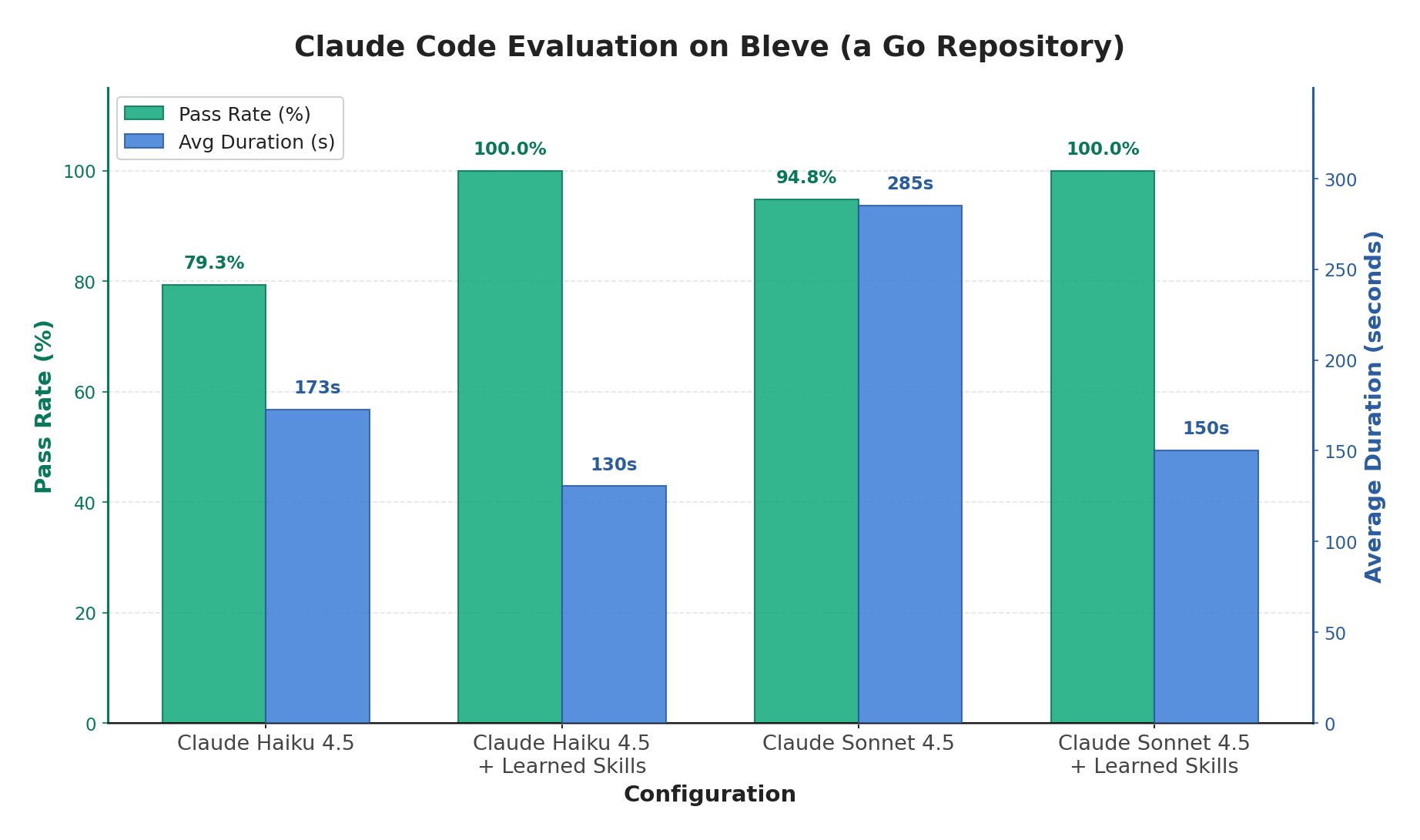}
  \caption{Claude Code on the Bleve repository. Optimized skills boost pass rates to near-perfect while reducing resolve time by 47\%. Skills transfer across models without reoptimization.}
  \Description{Bar chart showing pass rates: Haiku 4.5 79.3\% (173s), Haiku 4.5 + Skills 98.3\% (142s), Sonnet 4.5 94.8\% (285s), Sonnet 4.5 + Skills 100\% (169s).}
  \label{fig:skills}
\end{figure}

\textbf{Results.} Optimized skills boost Haiku 4.5's pass rate from 79.3\% to 98.3\% and Sonnet 4.5's from 94.8\% to 100\%, while cutting resolution time by 47\% (Figure~\ref{fig:skills}). Critically, skills discovered for one model transfer effectively to another without reoptimization, demonstrating the generalization mode's ability to learn model-agnostic repository knowledge.

\subsection{Cloud Scheduling Algorithms (Generalization)}\label{sec:cloud}

\textbf{Setup.} We optimize two cloud infrastructure algorithms from the ADRS benchmark~\cite{adrs2025}. \textbf{CloudCast} discovers broadcast routing strategies for multi-cloud data transfer, minimizing data egress cost. \textbf{Can't Be Late} learns scheduling policies deciding when to use cheap preemptible \texttt{SPOT} instances versus reliable \texttt{ON\_DEMAND} instances to meet deadlines. Both use generalization mode with training/validation splits over infrastructure scenarios.

\textbf{\asi{} design.} For CloudCast: per-partition routing decisions, edge utilizations, cost breakdowns. For Can't Be Late: spot-availability patterns, instance-usage timelines, segment counts (\texttt{SPOT} vs.\ \texttt{ON\_DEMAND} vs.\ restarts).

\begin{figure}[t]
  \centering
  \begin{subfigure}[b]{0.95\columnwidth}
    \includegraphics[width=\textwidth]{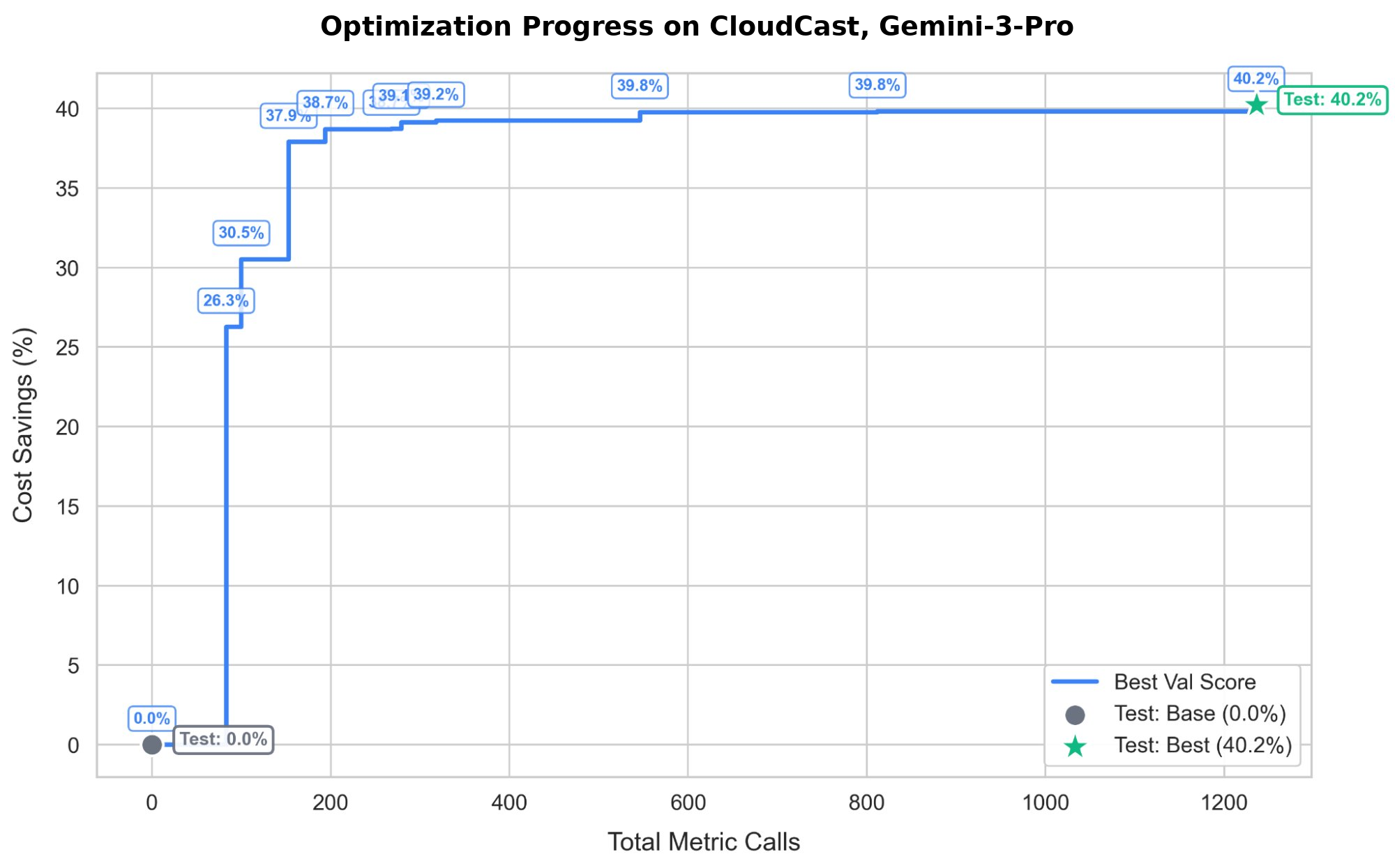}
    \caption{CloudCast: 40.2\% cost savings.}
    \label{fig:clouda}
  \end{subfigure}
  
  \begin{subfigure}[b]{0.95\columnwidth}
    \includegraphics[width=\textwidth]{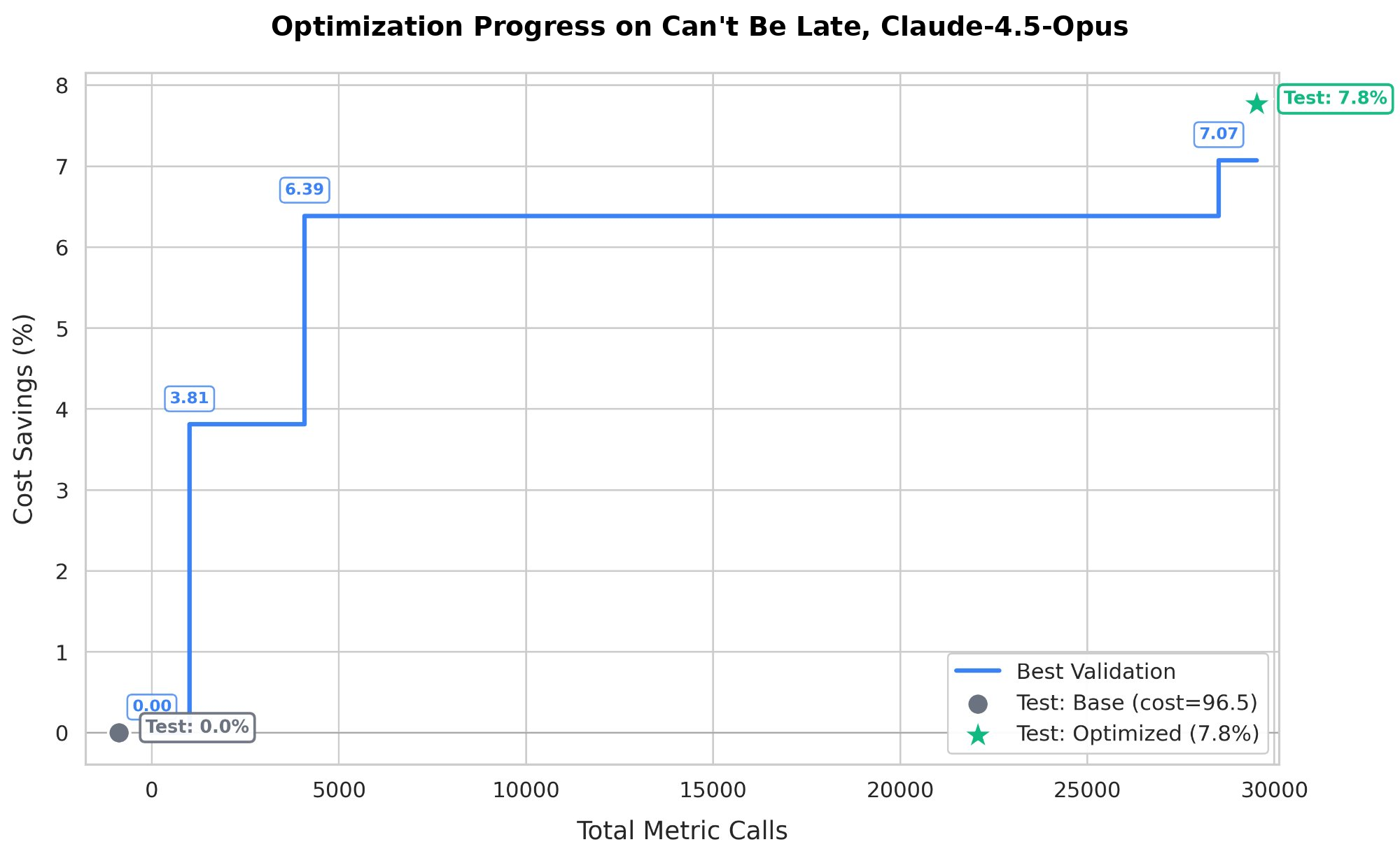}
    \caption{Can't Be Late: 7.8\% savings.}
    \label{fig:cloudb}
  \end{subfigure}
  \caption{Optimization trajectories for cloud scheduling. Both use generalization mode with train/val splits over infrastructure scenarios.}
  \Description{Two line charts showing optimization trajectories. CloudCast reaches 40.2\% test savings. Can't Be Late reaches 7.8\% test savings.}
  \label{fig:cloud}
\end{figure}

\textbf{Results.} CloudCast achieves 40.2\% cost savings over Dijkstra routing (Figure~\ref{fig:clouda}), evolving from a baseline shortest-path algorithm to a provider-aware Steiner tree approach that jointly optimizes for egress cost and transfer latency. Can't Be Late achieves 7.8\% cost savings (Figure~\ref{fig:cloudb}), evolving a simple deadline-check heuristic into an adaptive strategy with state tracking for spot-unavailability patterns, break-even switching cost analysis, and graduated decision thresholds based on slack ratio.
Both results top the ADRS leaderboard (\oa{}: 96.6 aggregate score vs.\ 92.9 for OpenEvolve, 72.0 for ShinkaEvolve). The evolved artifacts are qualitatively different from their seeds: CloudCast discovers provider-aware Steiner tree routing (absent from the Dijkstra seed), while Can't Be Late learns persistent spot-unavailability tracking and overhead-aware switching costs (absent from the greedy seed).

\subsection{ARC-AGI Agent Architecture (Generalization)}\label{sec:arc}
\textbf{Setup.} Rather than optimizing a prompt, we optimize the \emph{entire agent system}: code, sub-agent architecture, control flow, helper functions, and prompts are all treated as a single text artifact, building on an earlier proof-of-concept with GEPAAdapter~\cite{agrawal2025arc_agi_notebook}. The optimization objective is for the artifact to generalize to unseen ARC-AGI~\cite{chollet2019arc} puzzles.

\textbf{\asi{} design.} Training/test grid examples, per-puzzle scores, internal model outputs, LLM costs, error tracebacks, and code execution results.

\begin{figure}[t]
  \centering
  \includegraphics[width=\columnwidth]{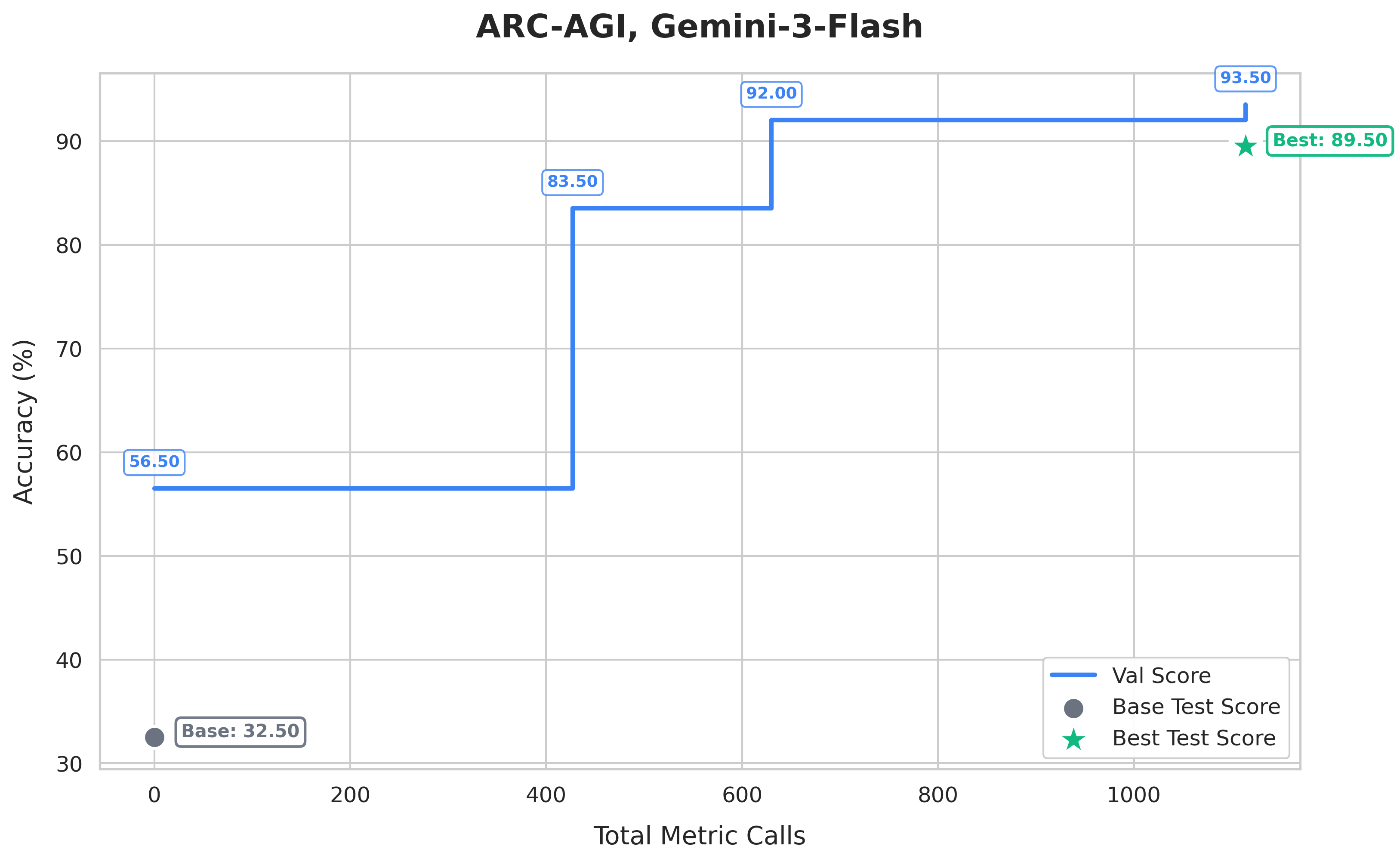}
  \caption{ARC-AGI agent architecture evolution with Gemini 3 Flash. Validation accuracy reaches 93.5\%; test accuracy improves from 32.5\% to 89.5\%.}
  \Description{Line chart showing validation accuracy improving from about 56\% to 93.5\% over metric calls. Base test 32.5\%, best test 89.5\%.}
  \label{fig:arc}
\end{figure}

\textbf{Results.} Using Gemini 3 Flash as both the proposer and the underlying agent model, \oa{} starts for a naive 10-line agent seed (one LLM call) and iteratively designs it into a 300+ line system consisting of 4 components along with fallbacks. The test accuracy improves from 32.5\% to \textbf{89.5\%}, a 57 percentage point gain (Figure~\ref{fig:arc}). The optimized architecture implements a 4-stage pipeline: (1) rule induction via pattern analysis, (2) code generation with \texttt{exec()}-based verification, (3) iterative debugging with up to 2 fix attempts, and (4) structured fallback from code-first to direct LLM prediction. This represents a qualitative leap: the system discovers architectural patterns (verify-then-fallback, iterative refinement) that typically require manual engineering iterations.

\subsection{AIME Prompt Optimization (Generalization)}\label{sec:aime}

\textbf{Setup.} We optimize a system prompt for GPT-4.1-mini on AIME (American Invitational Mathematics Examination) competition problems. Training uses AIME 2022--2024; testing uses AIME 2025.

\textbf{\asi{} design.} The evaluator returns each problem statement, the model's reasoning chain, extracted answer, ground truth, and a correct/incorrect flag.

\begin{figure}[t]
  \centering
  \includegraphics[width=\columnwidth]{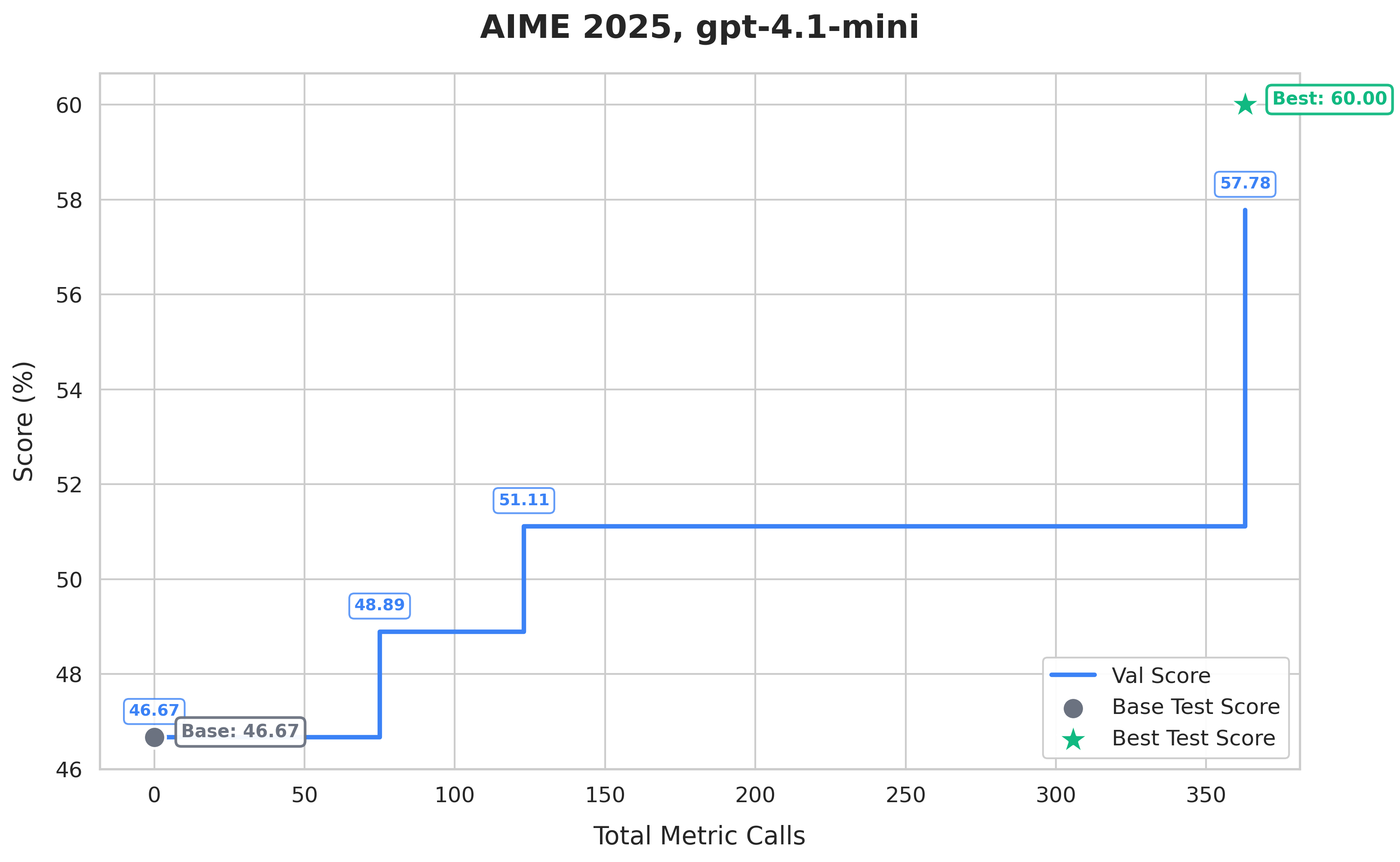}
  \caption{AIME prompt optimization for GPT-4.1-mini. Validation score improves from 46.67\% to 57.78\%; test score reaches 60.00\%.}
  \Description{Line chart showing validation score improving over 350 metric calls. Test accuracy reaches 60\% from 46.67\% baseline.}
  \label{fig:aime}
\end{figure}

\textbf{Results.} Prompt optimization improves GPT-4.1-mini from 46.67\% to \textbf{60.00\%} on AIME 2025 (Figure~\ref{fig:aime}), a 13.3pp gain from changing only the system prompt. This outperforms MIPROv2~\cite{opsahl2024miprov2} (51.33\% on the same benchmark). The optimized prompt (Appendix~\ref{app:sol-aime}) evolves from a single generic sentence into a structured 6-rule reasoning framework. This result matches the performance gains reported by~\citet{agrawal2026gepa}, demonstrating that exposing a prompt optimization algorithm through a general interface does not hurt performance on prompt optimization.

\subsection{CUDA Kernel Generation (Multi-Task Search)}\label{sec:cuda}

\textbf{Setup.} We generate CUDA kernels for 31 reference PyTorch operations from KernelBench~\cite{kernelbench2025}, evaluated on a V100 32GB GPU. The 31 problems span diverse operations: matrix multiplications, convolutions, reductions, element-wise ops, and normalization layers. Under the hood, \oa{} evolves the prompt that drives kernel generation; in multi-task mode, insights discovered for one problem (e.g., how to handle memory coalescing) transfer to others automatically through the shared Pareto frontier.

\textbf{\asi{} design.} The evaluator compiles the generated kernel, runs correctness tests (max absolute error vs. PyTorch reference), and benchmarks wall-clock time. \asi{} includes: (i) NVCC compiler errors with line numbers, (ii) correctness test failures with actual vs.\ expected outputs, (iii) relevant CUDA documentation snippets, and (iv) speedup ratio vs.\ the PyTorch baseline.

\begin{figure}[t]
  \centering
  \includegraphics[width=\columnwidth]{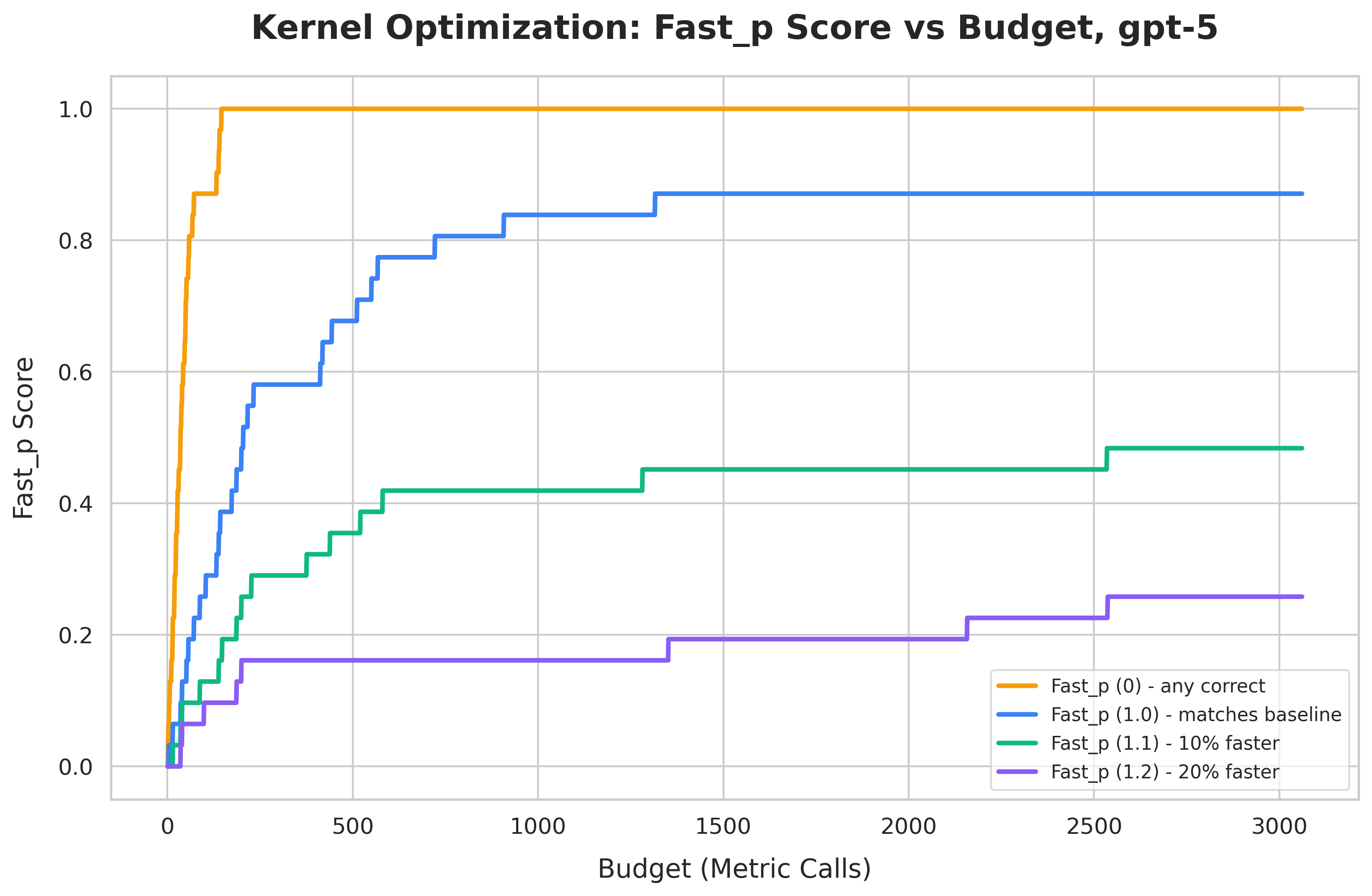}
  \caption{KernelBench results (GPT-5 as proposer). $\text{Fast}_p(s)$: fraction of kernels achieving speedup $\geq s$. 87\% match baseline; 25\% are 20\%+ faster.}
  \Description{Line chart showing Fast\_p at various speedup thresholds. Fast\_p(0)=100\%, Fast\_p(1.0)=87\%, Fast\_p(1.1)=48\%, Fast\_p(1.2)=25\%.}
  \label{fig:cuda-results}
\end{figure}

\textbf{Results.} 87\% of generated kernels match or beat the PyTorch baseline performance; 48\% achieve 10\%+ speedups, and 25\% achieve 20\%+ speedups (Figure~\ref{fig:cuda-results}). The evolved kernels employ techniques such as float4 vectorization, two-pass algorithms (compute statistics, then normalize), warp shuffle reductions, and shared memory tiling. Multi-task mode's advantages are analyzed in \S\ref{sec:multi-ablation}.

\subsection{Circle Packing (Single-Task Search)}\label{sec:circle}

\textbf{Setup.} The task is to pack $n{=}26$ circles while maximizing the sum of radii within a unit square. \oa{} optimizes the packing algorithm code; the evaluator executes the proposed packing code, and returns the score plus geometric diagnostics.

\textbf{\asi{} design.} Circle positions, radii, constraint violations, overlap distances, boundary violations, and a rendered visualization of the packing.

\begin{figure}[t]
  \centering
  \includegraphics[width=\columnwidth]{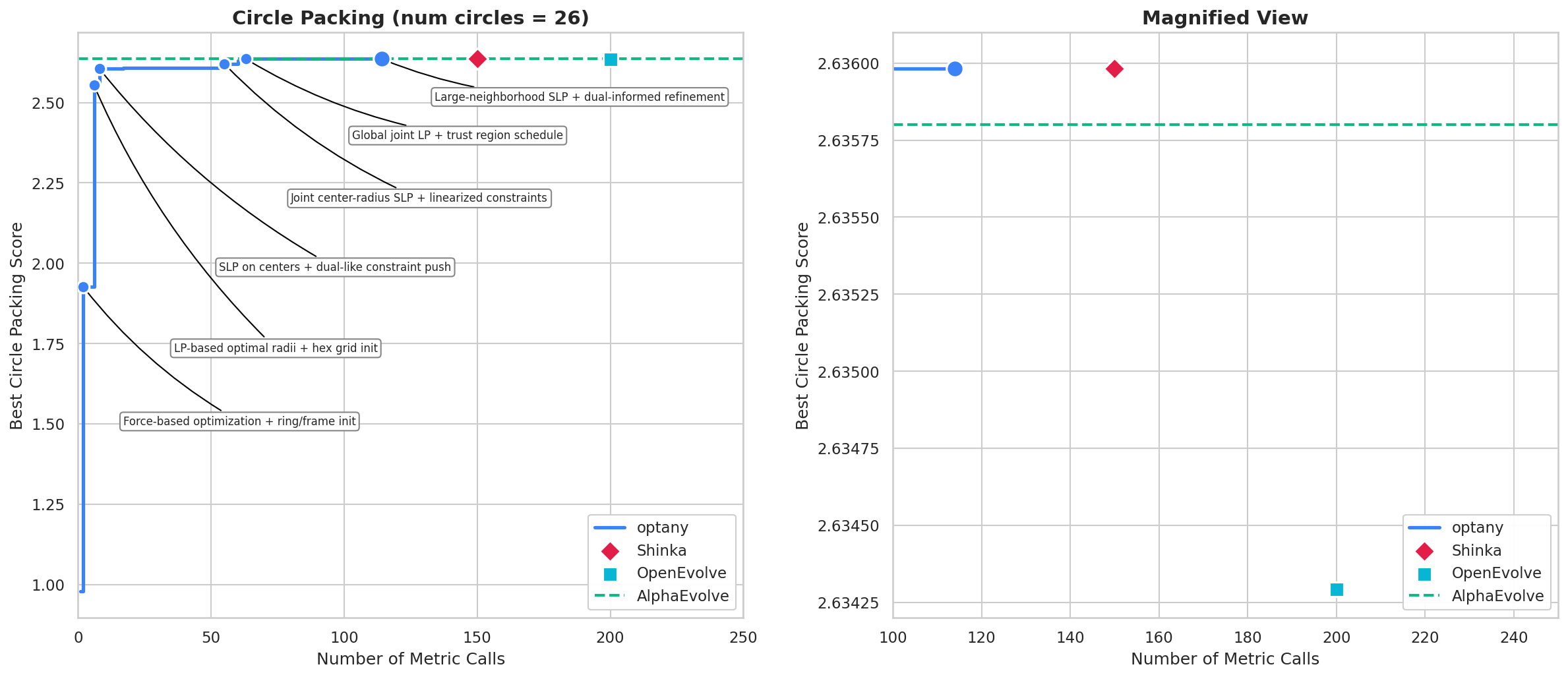}
  \caption{Circle packing ($n{=}26$). \oa{} outperforms AlphaEvolve's, ShinkaEvolve's, and OpenEvolve's solution, reaching a higher score with fewer evaluations.}
  \Description{Line chart comparing four methods on circle packing. \oa{} reaches highest score around 2.636 with fewer metric calls than alternatives.}
  \label{fig:circle}
\end{figure}

\textbf{Results.} \oa{} reaches a score of 2.63598+, outperforming AlphaEvolve's, OpenEvolve's, and ShinkaEvolve's reported solution (Figure~\ref{fig:circle}). The optimized algorithm is a bilevel optimizer: an LP over radii with dual-variable gradients for L-BFGS-B center optimization, augmented by CMA-ES exploration and diverse seeding strategies.

\paragraph{Controlled comparison with OpenEvolve.} To address concerns about comparing against published rather than reproduced results, we ran OpenEvolve (open-source reimplementation of AlphaEvolve) under matched conditions using the same proposer LLM (GPT-5.1). As shown in Table~\ref{tab:circle-controlled}, \oa{} achieved a superior score (2.63598) in just 63 evaluations (costing \~\$3.18), while OpenEvolve failed to match this performance even when given over three times the evaluation budget (200 iterations, costing \$6.85, reaching only 2.6307).

\begin{table}[t]
\caption{Controlled comparison of \oa{} vs.\ OpenEvolve on circle packing ($n{=}26$), both using GPT-5.1 as proposer.}
\label{tab:circle-controlled}
\small
\begin{tabular}{@{}lccc@{}}
\toprule
\textbf{Metric} & \textbf{\oa{}} & \textbf{OpenEv.@100} & \textbf{OpenEv.@200} \\
\midrule
Evaluations & 63 / 100 & 100 & 200 \\
Best sum\_radii & \textbf{2.63598} & 2.4583 & 2.6307 \\
Cost & \$3.18 & \$1.98 & \$6.85 \\
\bottomrule
\end{tabular}
\end{table}

\subsection{Image Generation (Multi-Task Search)}\label{sec:svg-gen}

\textbf{Setup.} We generate SVG code and CAD models (via \texttt{build123d}) for four image goals (Table~\ref{tab:optimization_examples} in Appendix~\ref{app:imggen-details}). The evaluator renders the image and queries a VLM to rate individual visual aspects on a 0--100 scale; each evaluator call scores one aspect, making this a natural multi-task search over the Pareto frontier of visual properties.

\textbf{Results.} Five human evaluators unanimously preferred\\ \oa{}-optimized images over zero-shot baselines across all goals. Quantitatively, the ``pelican riding a bicycle'' task achieves a VLM score of 0.726 vs.\ 0.330 for the zero-shot baseline (2.2$\times$ improvement). Qualitative comparisons are shown in Appendix Figure~\ref{fig:appendix_zero_shot_vs_oa}.

%% ================================================================
\subsection{Ablation: Multi-Task vs. Single-Task Search}\label{sec:multi-ablation}

\begin{figure}[t]
  \centering
  \includegraphics[width=\columnwidth]{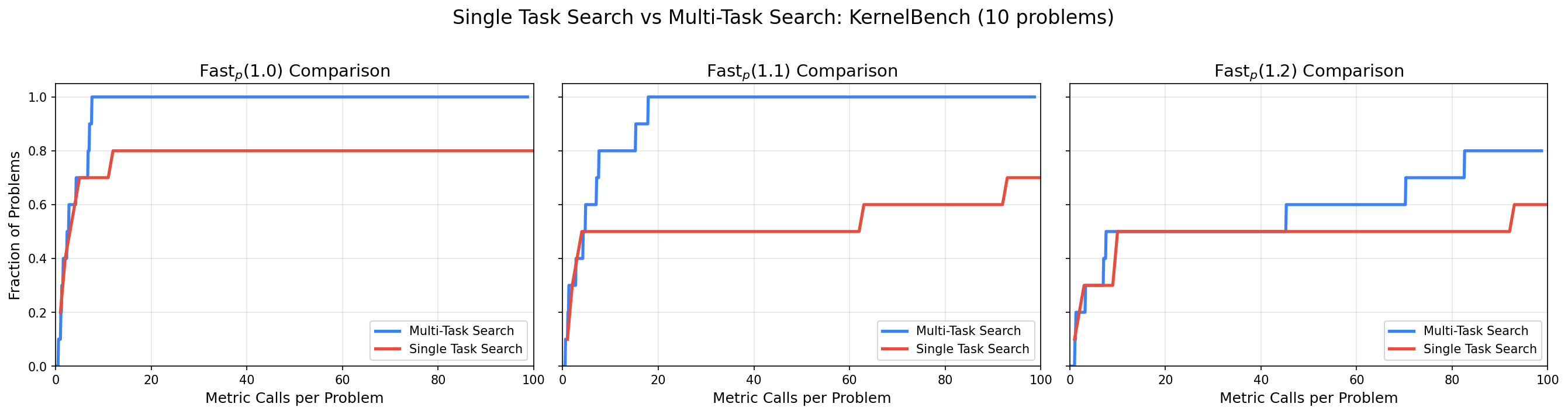}
  \caption{Single-task vs.\ multi-task mode on 10 selected KernelBench problems. Multi-task (blue) consistently outperforms single-task (red) at all speedup thresholds, converging faster and solving more problems.}
  \Description{Three line charts comparing single vs batch mode at F(1.0), F(1.1), F(1.2) thresholds. Batch mode solid lines are consistently above single mode dashed lines.}
  \label{fig:multi-ablation}
\end{figure}

We re-optimize the 10 best multi-task problems from scratch in single-task mode with equivalent per-problem budget. Figure~\ref{fig:multi-ablation} shows that multi-task mode consistently outperforms single-task across all speedup thresholds, with the gap widening at higher thresholds ($\text{Fast}_p(1.2)$: single-task plateaus early while multi-task continues improving).

The mechanism is cross-transfer via the Pareto frontier: optimization patterns discovered for one kernel (e.g., vectorized memory access, warp-level reductions) are preserved on the frontier and inform proposals for other kernels. In single-task mode, each problem must independently discover these patterns.

\paragraph{Scaling with number of tasks.} Multi-task benefits scale with the number of related tasks: MT20 (20 problems) outperforms MT10 (10 problems), which outperforms single-task, with gains most pronounced at moderate speedup thresholds (Tables~\ref{tab:mt-scaling}--\ref{tab:mt-scaling-20} in Appendix~\ref{app:mt-scaling}). Frontier size does not bottleneck scaling, as candidates are sampled by frontier frequency (e.g., ARC-AGI used 200 tasks effectively).

\subsection{Ablation: Side Information}\label{sec:asi-ablation}

To isolate the contribution of \asi{}, we compare \oa{} with and without actionable side information (sub-scores) on prompt optimization for the Facility Support Analysis dataset. In the ``with \asi{}'' condition, the evaluator returns per-aspect sub-scores alongside the aggregate score. In the ``without \asi{}'' condition, only the aggregate score is returned.

\begin{figure}[t]
  \centering
  \includegraphics[width=\columnwidth]{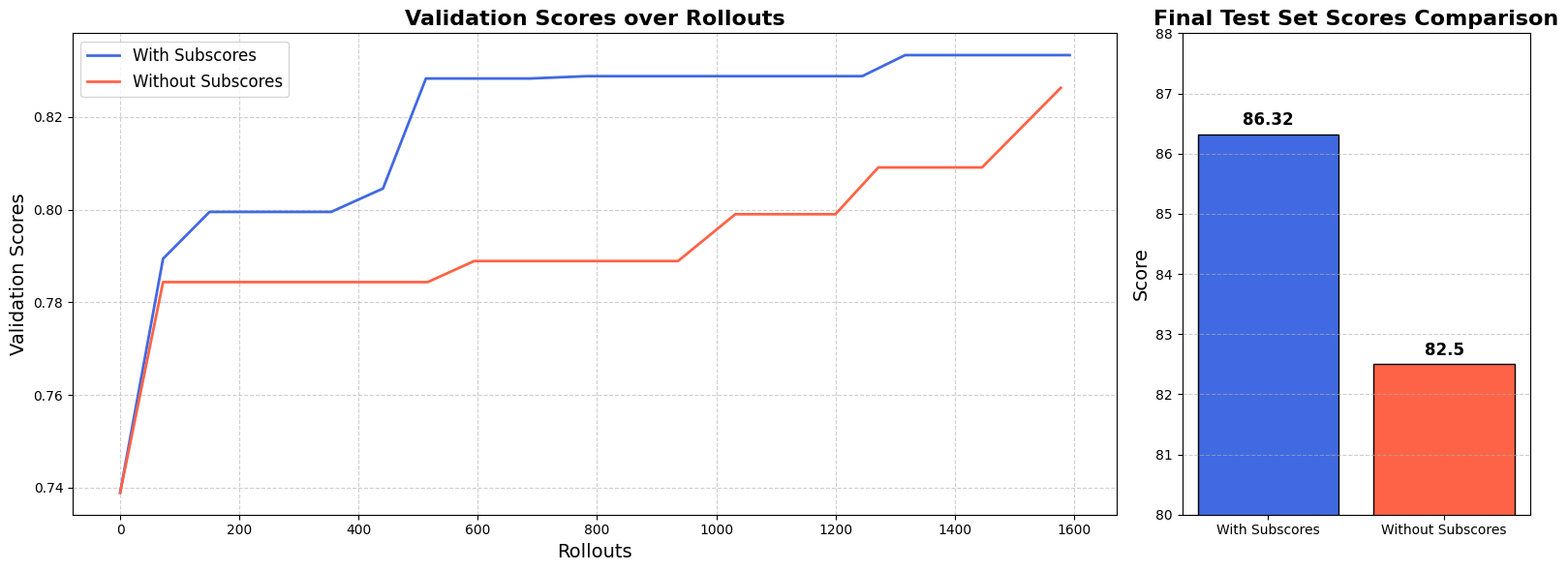}
  \caption{Ablation: prompt optimization with vs.\ without \asi{} on the Facility Support Analysis dataset. \asi{} accelerates convergence (left) and improves final test performance (right): 86.32 vs.\ 82.5.}
  \Description{Left: validation score curves showing with-subscores (blue) converging faster than without (red). Right: bar chart showing final test scores 86.32 vs 82.5.}
  \label{fig:asi-ablation}
\end{figure}

Figure~\ref{fig:asi-ablation} shows two effects. First, \asi{} accelerates convergence: the ``with \asi{}'' condition reaches a validation score of 0.80 within 100 rollouts, while the score-only condition requires approximately 600 rollouts to reach the same level. Second, \asi{} improves final performance: the test score with \asi{} is 86.32 versus 82.5 without.

Sub-scores let the proposer identify which aspects are strong vs.\ weak and target revisions accordingly, rather than receiving only an aggregate signal.

\paragraph{Cross-domain SI ablation.} SI vs.\ score-only ablations on circle packing and CUDA kernels (Table~\ref{tab:si-cross-domain}) confirm generalization: SI achieves the optimal circle packing solution (score-only reaches 94\%), and enables 2.5--5$\times$ more kernels to exceed speedup thresholds. SI reveals \emph{which} failure mode to address next; without it, the proposer can only observe that the score changed.

\begin{table}[t]
\caption{SI vs.\ score-only ablation across three domains. SI provides substantial gains in all domains, confirming generalization beyond prompt optimization.}
\label{tab:si-cross-domain}
\small
\begin{tabular}{@{}llcc@{}}
\toprule
\textbf{Domain} & \textbf{Metric} & \textbf{With SI} & \textbf{Score only} \\
\midrule
Circle Packing & Best score (\% of best) & \textbf{100\%} & 93.96\% \\
KernelBench (ST) & Kernels $\geq$1.1$\times$ ($f_{1.1}$) & \textbf{32.3\%} & 12.9\% \\
KernelBench (ST) & Mean speedup & \textbf{4.11$\times$} & 1.15$\times$ \\
KernelBench (MT) & Kernels $\geq$1.1$\times$ ($f_{1.1}$) & \textbf{40\%} & 0\% \\
KernelBench (MT) & Mean speedup & \textbf{1.15$\times$} & 1.03$\times$ \\
\bottomrule
\end{tabular}
\end{table}

\subsection{Proposer Sensitivity and Optimization Cost}\label{sec:proposer-sensitivity}%
\label{sec:cost}

Comparing GPT-5.1 against the cheaper GPT-5-nano reveals a clear cost-performance tradeoff (Table~\ref{tab:proposer-sensitivity} in Appendix~\ref{app:proposer-cost}): the nano model reduces costs by over 90\% on Circle Packing while still improving substantially over the seed, but consistently underperforms the larger model on final quality. Total optimization costs range from \$1 (Numerical Blackbox) to \$144.70 (ARC-AGI), with reflection cost minimal and total spend dominated by the evaluator (Table~\ref{tab:optimization-cost} in Appendix~\ref{app:proposer-cost}).

%% ================================================================
\section{Why the Framework Works: Optimization Trajectory Analysis}\label{sec:trajectory}

Beyond final scores, trajectory analysis on circle packing reveals three key mechanisms driving \oa{}'s effectiveness (detailed in Appendix~\ref{app:trajectory-detail}):
(1)~\textbf{SI enables targeted algorithmic shifts}: SI reveals \emph{which} failure mode to address next (e.g., collapsed radii $\to$ switch to LP; poor centers $\to$ switch to SLP), enabling directed rather than blind mutations.
(2)~\textbf{Multi-module Pareto leapfrogging}: the code artifact and refiner prompt are both tracked on the Pareto front; each module's advances become the foundation for the other's next improvement, creating a productive coordination dynamic absent from single-artifact systems.
(3)~\textbf{Pareto diversity prevents premature convergence}: the front retains candidates from multiple algorithmic families (greedy, LP, SLP, bilevel L-BFGS, CMA-ES), ensuring structurally diverse parents for proposals.
These mechanisms operate identically across domains because they arise from the \texttt{evaluate(candidate)} $\to$ \texttt{(score, side\_info)} contract.

%% ================================================================
\section{Discussion}\label{sec:discussion}

\paragraph{When does multi-task search help?}
Our experiments reveal that cross-task transfer is most beneficial when problems share underlying optimization patterns but differ in their specifics. CUDA kernel generation exemplifies this: memory coalescing, vectorized access patterns, and warp-level reductions are strategies that apply across operations but manifest differently for each kernel. Multi-task mode discovers these patterns once and transfers them, while single-task mode must rediscover them independently for each problem (Tables~\ref{tab:mt-scaling}--\ref{tab:mt-scaling-20}).

\paragraph{When does multi-task search hurt?}
Multi-task search can degrade performance when tasks lack shared transferable structure. We quantify this on circle packing, where optimizing different values of $N$ jointly introduces noise rather than useful cross-transfer:

\begin{table}[t]
\caption{Multi-task search on circle packing. Unlike CUDA kernels, circle packing problems for different $N$ are fundamentally independent, and multi-task search introduces noise.}
\label{tab:mt-circle-hurt}
\small
\begin{tabular}{@{}lc@{}}
\toprule
\textbf{Config} & \textbf{Score} \\
\midrule
Single-task & \textbf{2.6360} \\
MT7  & 2.6313 \\
MT11 & 2.5973 \\
\bottomrule
\end{tabular}
\end{table}

Circle packing problems for different $N$ are fundamentally independent, optimal configurations change unpredictably with $N$, with no transferable structure~\cite{graham1996dense,galiev2013linear}. In general, multi-task search helps when tasks share underlying patterns (e.g., CUDA kernels on the same hardware) and hurts when they are fundamentally independent.

\paragraph{The role of \asi{} across domains.}
While the SI ablation (Table~\ref{tab:si-cross-domain}) confirms SI's value, the mechanism differs by domain: for code (CUDA, circle packing), SI surfaces compiler errors and runtime diagnostics pinpointing failures; for agents (ARC-AGI), per-puzzle traces reveal which components fail; for cloud scheduling, SI exposes temporal decision structure. In each case, SI converts a scalar signal into actionable diagnostics.

\paragraph{Artifacts optimized by \oa{}.}
The optimized artifacts range from structured prompts (AIME) and agent architectures (ARC-AGI) to 900+ line bilevel algorithms (circle packing), demonstrating that the system discovers qualitatively novel strategies---multi-stage pipelines (ARC-AGI), provider-aware Steiner trees (CloudCast), break-even cost analysis (Can't Be Late)---arising from the interaction between LLM reasoning and diagnostic feedback.

%% ================================================================
\section{Limitations}\label{sec:limitations}

\oa{} inherits limitations from LLM-based optimization. (1) The quality of proposals depends on the proposer LLM's capabilities; weaker models produce weaker candidates, as confirmed by our proposer sensitivity analysis (Table~\ref{tab:proposer-sensitivity}). (2) Evaluation cost can be high when the evaluator involves expensive operations (e.g., \$144 for ARC-AGI, Table~\ref{tab:optimization-cost}), however, it must be noted that LLM-based optimization is highly sample efficient and therefore calls evaluators less often. (3) The system assumes the artifact is representable as text; optimization of continuous parameters or binary artifacts requires a text-based proxy. (4) While multi-task search provides cross-transfer benefits on related problems, the degree of benefit depends on how related the problems are, for example, circle packing exhibits degradation with multi-task mode (Table~\ref{tab:mt-circle-hurt}). (5) designing effective SI still requires domain expertise; while evaluators returning only a score work, the demonstrated gains come from expert-designed SI (compiler errors, profiler traces, VLM scoring rubrics). That said, \oa{} trades \emph{optimization} expertise for \emph{domain} expertise. The user, most often a domain expert, need not configure backends, tune algorithmic hyperparameters, or engineer prompting strategies, only surface the diagnostics they already understand.

%% ================================================================
\section{Conclusion}\label{sec:conclusion}

\oa{} demonstrates that a simple declarative interface (seed artifact, evaluator, and optional dataset) is sufficient to match or outperform purpose-built tools across diverse domains. The key ideas are (1) three unified optimization modes under one API, (2) Side Information as a first-class evaluator contract, and (3) Pareto-based search across metrics and examples. The API is backend-agnostic; as new optimization strategies emerge, they plug in without changing user code. \oa{} is open-sourced with multiple backends as a part of the GEPA project: \url{https://github.com/gepa-ai/gepa}.

%% ================================================================
\begin{acks}
This research is supported in part by gifts from Accenture, Amazon, AMD, Anyscale, Broadcom, Google, IBM, Intel, Intesa Sanpaolo, Lambda, Lightspeed, Mibura, NVIDIA, Samsung SDS, SAP, by the U.S. Department of Energy, Office of Science, Office of Advanced Scientific Computing Research through the X-STACK: Programming Environments for Scientific Computing program (DESC0021982), and the Defense Advanced Research Projects Agency (DARPA) under Agreement No.\ HR00112590134. Lakshya A Agrawal is supported by a Laude Slingshot grant provided by the Laude Institute and an Amazon AI PhD Fellowship.
\end{acks}

\bibliographystyle{ACM-Reference-Format}
\bibliography{references}

%% ================================================================
\newpage

\appendix

\section{Use of Generative AI}
The authors made use of Generative AI technologies including ChatGPT, Gemini, Claude and Cursor to generate sections of this Work, including text, tables, graphs, code, etc. The experiment design and details were explicitly the authors' original ideas.

\section{Blackbox Mathematical Optimization}\label{app:optuna}

We additionally evaluate \oa{} in single-task search mode on blackbox mathematical optimization, using the 56-problem EvalSet benchmark~\cite{evalset2018} against Optuna~\cite{akiba2019optuna}. Rather than tuning parameters within a fixed algorithm, \oa{} optimizes the solver code itself, discovering bespoke algorithms for each problem.

With a budget of 8,000 evaluations per problem, \oa{} ties Optuna on 40 problems, wins 7, and loses 9. On 10 selected problems where Optuna struggles with lower budgets (2,000 evaluations), \oa{} finds better solutions on 7 out of 10. The mechanism: Optuna's fixed TPE-CMA-ES pipeline fails in predictable, structural ways (e.g., TPE's per-dimension sampling converges to trap basins; CMA-ES assumes smooth unimodal landscapes). \oa{} tailors the solver to each problem---discovering L-BFGS-B for boundary optima and multi-start search for deceptive traps.

\section{Seedless Mode: 3D Unicorn}\label{app:unicorn}

Every main experiment starts from a seed artifact. Seedless mode (\texttt{seed\_candidate=None}) instead provides only a natural-language objective and lets the LLM bootstrap the first candidate. We demonstrate this on a 3D modeling task: generating a Python script (build123d + pyrender) that produces a 3D unicorn. The evaluator renders multi-view PNGs and asks a VLM to score them, passing images back as \asi{}. Starting from no code, \oa{} iteratively refines geometry, proportions, and anatomical detail, producing a recognizable 3D unicorn that improves substantially over the zero-shot baseline.

\section{Detailed Algorithm}\label{app:algorithm}

\begin{algorithm}[H]
\caption{\oa{}: Core optimization loop}
\label{alg:oa}
\begin{algorithmic}[1]
\small
\REQUIRE Artifact $\Phi_0$, evaluator $f$, dataset $\mathcal{D}$, budget $B$
\REQUIRE Minibatch size $b$, Pareto set size $n$
\STATE Initialize candidates $\mathcal{P} \gets [\Phi_0]$
\STATE Evaluate $\Phi_0$ on $\mathcal{D}$; record per-example scores $S$
\WHILE{budget $B$ not exhausted}
  \STATE $k \gets$ \textsc{ParetoSelect}($\mathcal{P}, S$) \COMMENT{Select based on frontier}
  \STATE $\mathcal{M} \gets$ minibatch of size $b$ from $\mathcal{D}$
  \STATE Execute $\Phi_k$ on $\mathcal{M}$; collect scores and \asi{}
  \STATE $\Phi' \gets$ \textsc{Reflect}($\Phi_k$, scores, \asi{}) \COMMENT{LLM proposes fix}
  \IF{$\Phi'$ improves on $\mathcal{M}$}
    \STATE Evaluate $\Phi'$ on full $\mathcal{D}$
    \STATE $\mathcal{P} \gets \mathcal{P} \cup \{\Phi'\}$
    \STATE Update $S$; prune dominated candidates
  \ENDIF
\ENDWHILE
\STATE \textbf{return} $\Phi^* \in \mathcal{P}$ maximizing average score
\end{algorithmic}
\end{algorithm}

Algorithm~\ref{alg:oa} presents the core loop. For single-task search, the ``dataset'' is a singleton and per-example tracking reduces to per-metric tracking. For multi-task search, each dataset element is an independent problem. For generalization, scores on $\mathcal{D}$ guide search while a held-out \texttt{valset} measures generalization. The \textsc{ParetoSelect} subroutine shows the candidate selection algorithm used in the default optimization backend, GEPA~\cite{agrawal2026gepa} which identifies non-dominated candidates and samples proportionally to their frontier frequency.

\section{Multi-Task Scaling Tables}\label{app:mt-scaling}

\begin{table}[h]
\caption{Multi-task scaling on 10 KernelBench problems. $f_{1.x}$: fraction of kernels achieving $\geq$$x$\% speedup over PyTorch baseline.}
\label{tab:mt-scaling}
\small
\begin{tabular}{@{}lccc@{}}
\toprule
\textbf{Setting} & $\mathbf{f_{1.0}}$ & $\mathbf{f_{1.1}}$ & $\mathbf{f_{1.2}}$ \\
\midrule
ST   & 60.0\% & 40.0\% & \textbf{20.0\%} \\
MT10 & \textbf{90.0\%} & 40.0\% & \textbf{20.0\%} \\
MT20 & \textbf{90.0\%} & \textbf{50.0\%} & \textbf{20.0\%} \\
\bottomrule
\end{tabular}
\end{table}

\begin{table}[h]
\caption{Single-task vs.\ MT20 on 20 randomly sampled KernelBench problems.}
\label{tab:mt-scaling-20}
\small
\begin{tabular}{@{}lccc@{}}
\toprule
\textbf{Setting} & $\mathbf{f_{1.0}}$ & $\mathbf{f_{1.1}}$ & $\mathbf{f_{1.2}}$ \\
\midrule
ST   & 50.0\% & 25.0\% & \textbf{15.0\%} \\
MT20 & \textbf{90.0\%} & \textbf{40.0\%} & \textbf{15.0\%} \\
\bottomrule
\end{tabular}
\end{table}

\section{Optimization Trajectory Analysis: Full Details}\label{app:trajectory-detail}

\paragraph{Mechanism 1: SI enables targeted algorithmic shifts.}
SI works because it reveals \emph{which} failure mode to address next, not merely that performance changed. In circle packing, SI-driven reflection produces a characteristic pattern: collapsed radii $\to$ switch to LP; poor center placement $\to$ switch to SLP; local saturation $\to$ switch to bilevel L-BFGS. Without SI, the proposer can only observe that the score changed, not why, and resorts to undirected mutations. The cross-domain SI ablation (Table~\ref{tab:si-cross-domain}) confirms this mechanism generalizes: on KernelBench with multi-task search, SI enables 40\% of kernels to exceed 1.1$\times$ speedup vs.\ 0\% with score-only feedback.

\paragraph{Mechanism 2: Multi-module Pareto leapfrogging.}
\oa{} optimizes both the code artifact and a refiner prompt, both tracked on the shared Pareto front. In circle packing, this creates a productive leapfrogging dynamic: the refiner discovers LP-based optimization while the code module is still a weak heuristic (code=0.98, refiner=1.93). The code module then absorbs the LP approach, catching up ($\to$2.61). The refiner pushes further with SLP ($\to$2.63). The code module absorbs SLP and reaches the world record. Each module's advances become the foundation for the other's next improvement---a coordination mechanism absent from single-artifact systems like AlphaEvolve. Even broken code mutations (score=0.0) are recovered by the refiner and retained on the front, acting as a safety net that preserves exploration.

\paragraph{Mechanism 3: Pareto diversity prevents premature convergence.}
At convergence, the Pareto front retains candidates from multiple algorithmic families simultaneously (greedy, LP, SLP, bilevel L-BFGS, CMA-ES) across quality dimensions (max score, mean score, EMA stability, improvement rate). This ensures the proposer has access to structurally diverse parents when generating new candidates, rather than being locked into refining a single approach. The preservation of diverse strategies is what enables the algorithmic shifts described above: even when LP dominates on raw score, greedy and CMA-ES candidates survive on stability metrics and can seed novel hybrid approaches.

\section{Proposer Sensitivity and Optimization Cost}\label{app:proposer-cost}

\begin{table}[h]
\caption{Proposer LLM sensitivity. GPT-5-nano reduces cost significantly but underperforms GPT-5.1 on final achieved performance. Both models improve substantially over the seed.}
\label{tab:proposer-sensitivity}
\small
\begin{tabular}{@{}llcc@{}}
\toprule
\textbf{Model} & \textbf{Task} & \textbf{Performance} & \textbf{Cost} \\
\midrule
GPT-5.1 & AIME & 46.67\% $\to$ \textbf{60.0\%} & \$6.44 \\
GPT-5-nano & AIME & 46.67\% $\to$ 50.0\% & \$3.71 \\
\midrule
GPT-5.1 & Circle Pack. & 0.98 $\to$ \textbf{2.636} & \$6.00 \\
GPT-5-nano & Circle Pack. & 0.98 $\to$ 2.512 & \$0.50 \\
\bottomrule
\end{tabular}
\end{table}

\begin{table}[h]
\caption{Total optimization cost per experiment. Reflection cost is minimal; total spend is dominated by the evaluator.}
\label{tab:optimization-cost}
\small
\begin{tabular}{@{}ll@{}}
\toprule
\textbf{Task} & \textbf{Total Cost} \\
\midrule
Numerical Blackbox & \$1 \\
Circle Packing & \$6 \\
KernelBench (31 kernels) & \$140 (\$4.51/kernel $\times$ 31) \\
AIME (Generalization) & \$6.44 (\$2.17 reflection, \$4.27 LLM) \\
SVG Optimization & \$18 \\
Agent Skills (Generalization) & \$50 \\
Cloud Scheduling & \$52.42 \\
ARC-AGI (Generalization) & \$144.70 (\$0.70 reflection, \$144 agent) \\
\bottomrule
\end{tabular}
\end{table}

\section{Image Generation Details}\label{app:imggen-details}

\begin{table}[h]
    \centering
    \small
    \begin{tabular}{@{}p{3.5cm}ccl@{}}
    \toprule
    \textbf{Goal} & \textbf{\# Aspects} & \textbf{Repr.} & \textbf{Model} \\
    \midrule
    A pelican riding a bicycle & 12 & SVG & Gemini Flash 3.0 \\
    A high-quality 3D unicorn & 4 & CAD & Claude Opus 4.6 \\
    An octopus on a pipe organ & 13 & SVG & Claude Opus 4.6 \\
    A sloth steering an excavator & 13 & SVG & Claude Opus 4.6 \\
    \bottomrule
    \end{tabular}
    \caption{Image generation goals. A VLM evaluator scores one visual aspect per call; multi-task search explores the Pareto frontier of visual properties.}
    \label{tab:optimization_examples}
\end{table}

For SVG tasks, the evaluator renders the image and queries a VLM for feedback. For each goal, we define several natural language properties which ask a VLM to rate on a scale of 0 to 100 how well the image aligns with that aspect. During each evaluator call, the VLM rates one aspect (not all at once), making this a natural multi-task search over the Pareto frontier. In the CAD setting, since we are dealing with 3D objects, the evaluator takes 3 screenshots equidistant apart and asks the VLM to provide feedback using those images.

\section{Optimized AIME Prompt}\label{app:sol-aime}

\begin{promptbox}[Optimized Prompt for AIME]
Solve the math problem carefully and thoroughly. Your goal is to produce a correct,
well‑structured solution that leads unambiguously to the requested final result.

Follow these rules:

1. Restate the problem briefly in your own words.

2. Set up notation and equations cleanly before manipulating them.
   - Define variables explicitly.
   - State all constraints (e.g., integrality, ranges, geometric conditions) before
     using them.

3. Show clear, logically ordered reasoning.
   - Justify each important algebraic or geometric step.
   - When you split into cases, state why each case is necessary and what assumptions
     define it.
   - If you invoke a known theorem (e.g., Ptolemy, Power of a Point, similarity, Vieta),
     name it and show exactly how it applies in this context.

4. Handle dead ends correctly.
   - If you realize a line of reasoning leads to a contradiction or dead end, explicitly
     say so.
   - Then restart from the last correct point; do not guess or hand‑wave.
   
5. Keep the reasoning focused and minimal while still being rigorous.
   - Avoid unnecessary numerical approximations if an exact approach is available.
   - Do not approximate exact values unless the problem explicitly asks for a decimal.
   - Prefer algebraic or structural arguments over trial‑and‑error or random guessing.
   - You may test candidate values only after deriving strong constraints that sharply
     limit the possibilities.

6. At the end, clearly isolate the answer:
   - Provide the final answer as a single number or expression on its own line.
   - Do not include any extra words, symbols, or explanation on that final line.
\end{promptbox}

\section{Discovered solutions}
\label{app:discovered_solutions}

We present excerpts of the final optimized artifacts discovered by \oa{} for each domain. 

\subsection{Coding Agent Skills: Bleve Repository}
\label{app:sol-skills}

The following is the optimized \texttt{SKILL.MD} excerpt discovered by \oa{} for the Bleve search library:

\begin{promptbox}[Optimized Bleve Skills (excerpt)]
4) Run tests early and iterate from failures (tests are the bug report)
- Start broad when feasible: `cd /testbed \&\& go test ./...` (or project equivalent).
- Narrow quickly:
  - package: `go test ./path/to/pkg`
  - single test: `go test ./path/to/pkg -run TestName -count=1` (add -v only if needed)
- For panics: follow the stack trace top frame in repo code first.
- For mismatches: use “expected vs got” to locate the producing function and invariants.

...

7) Make minimal, reviewable changes and verify continuously
- Change one behavior at a time; rerun the smallest reproducing test after each change.
- Add focused unit tests when coverage is missing; keep them in the same package and table-driven where sensible (include short words + accented/Unicode edge cases).
- Avoid scratch main.go files in repo root.
\end{promptbox}

\subsection{ARC-AGI Agent Architecture}
\label{app:sol-arc}

The optimized agent grew from a 10-line seed to a 300+ line system implementing a 4-stage pipeline: rule induction via pattern analysis, code generation with \texttt{exec()}-based verification, iterative debugging with up to 2 fix attempts, and structured fallback from code-first to direct LLM prediction.

\begin{figure}[h]
  \centering
  \includegraphics[width=\columnwidth]{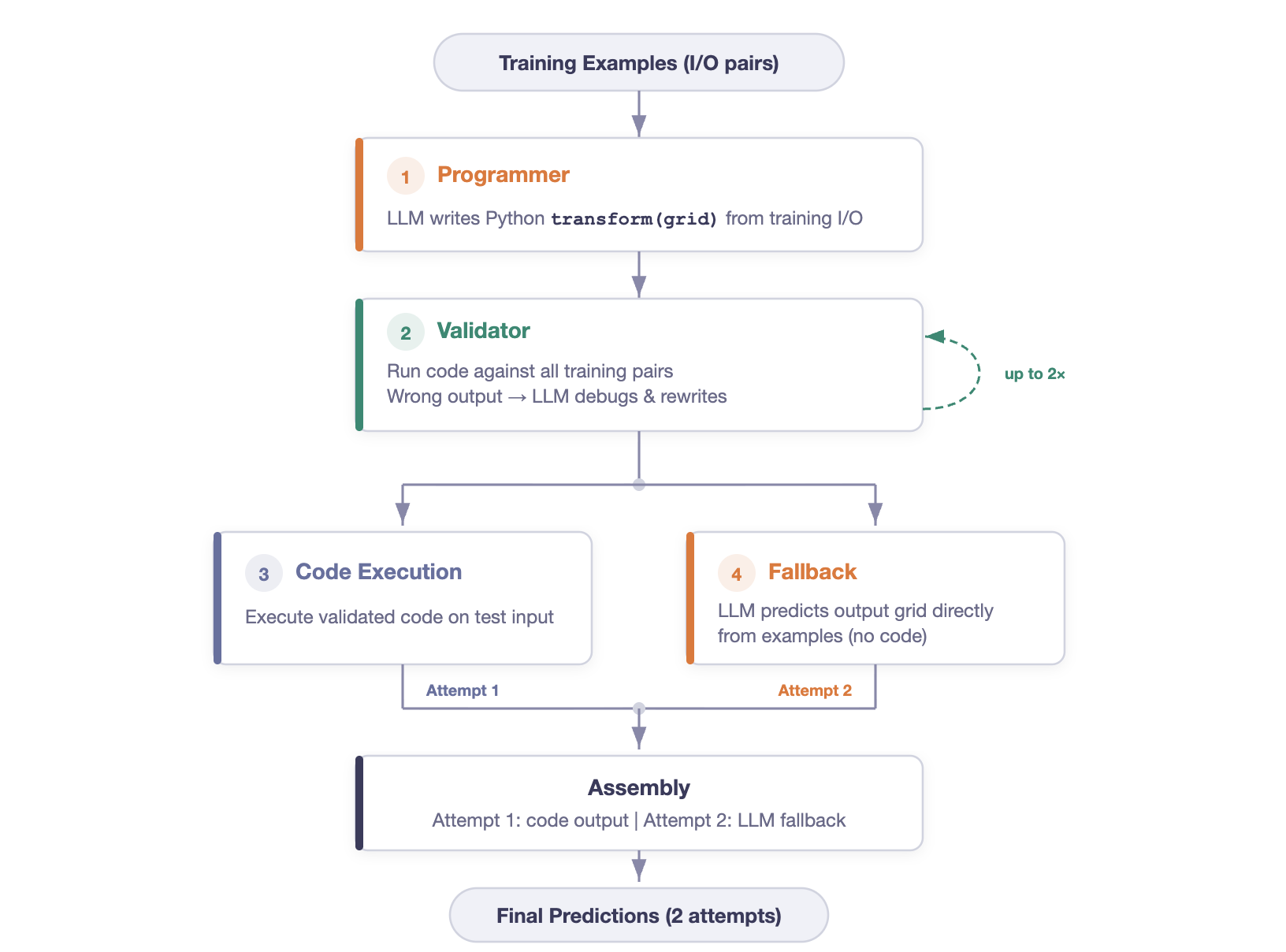}
  \caption{Architecture of the optimized ARC-AGI agent. The system discovers a 4-stage pipeline with verify-then-fallback logic, starting from a naive single-call seed.}
  \Description{Architecture diagram of the optimized ARC-AGI agent showing four stages: rule induction, code generation with exec()-based verification, iterative debugging, and structured fallback.}
  \label{fig:arc-architecture}
\end{figure}

\subsection{CloudCast Routing Algorithm}
\label{app:sol-cloudcast}

The optimized CloudCast algorithm (178 lines) discovers provider-aware Steiner tree routing with egress cost optimization, a qualitative departure from the Dijkstra seed. 

We show the main \texttt{search\_algorithm} function; the full artifact is available in the supplementary material.

\begin{promptbox}[Optimized CloudCast Algorithm (excerpt)]
\begin{lstlisting}[aboveskip=0pt,belowskip=0pt,numbers=none,frame=none,backgroundcolor=\color{gray!5}]
def search_algorithm(src, dsts, G, num_partitions):
    """Optimized Broadcast Routing Algorithm v3.
    Key Optimizations:
    1. Provider-Aware Weighting: biases path finding towards
       intra-provider links to minimize egress.
    2. Pareto-Frontier Candidate Selection: Explicitly keeps
       candidates that offer distinct cost/time tradeoffs.
    3. Diverse Steiner Strategies: Includes MST-like
       approximations for cost and bottleneck-widest
       paths for throughput.
    4. Robust Greedy Allocation: Accurately models bandwidth
       contention across partitions.
    """
    # --- Constants & Configuration ---
    EST_DATA_VOL_GB = 300.0
    EST_INSTANCE_COST_PER_HR = 10.0
    PARTITION_VOL_GB = EST_DATA_VOL_GB / max(1, num_partitions)
    # Sweep parameters for Cost vs Time tradeoff
    alphas = [0.0, 1e-5, 0.001, 0.01, 0.05, 0.1, 0.5, 2.0]
    bw_thresholds = [0.0, 0.5, 5.0, 20.0]
    strategies = ['prim', 'prim', 'furthest', 'random']
    # ... (remaining 140 lines: provider extraction, graph
    #  preprocessing, Steiner tree construction, greedy
    #  allocation, and Pareto-frontier selection)
\end{lstlisting}
\end{promptbox}

\subsection{Can't Be Late Scheduling Policy}
\label{app:sol-cantbelate}

The optimized scheduling policy (110 lines) starts from a simple deadline-check heuristic and discovers three key behaviors absent from the seed: (1) break-even switching cost analysis that avoids costly SPOT$\to$ON\_DEMAND transitions when remaining work is small, (2) persistent spot-unavailability tracking via a counter that detects when SPOT is unlikely to return, and (3) graduated decision thresholds based on slack ratio that become increasingly aggressive as the deadline approaches. We show the core \texttt{\_step} method; the full artifact includes \texttt{reset()} and additional edge-case guards.

\begin{promptbox}[Optimized Can't Be Late Policy (excerpt)]
\begin{lstlisting}[aboveskip=0pt,belowskip=0pt,numbers=none,frame=none,backgroundcolor=\color{gray!5}]
from sky_spot.strategies.strategy import Strategy
from sky_spot.utils import ClusterType

class EvolveSingleRegionStrategy(Strategy):
    def __init__(self, args):
        super().__init__(args)
        self.spot_unavailable_count = 0
        self.consecutive_short_spot_windows = 0

    def _step(self, last_cluster_type, has_spot) -> ClusterType:
        remaining_task_time = self.task_duration - sum(self.task_done_time)
        remaining_time = self.deadline - self.env.elapsed_seconds
        slack = remaining_time - remaining_task_time - self.restart_overhead

        # Track persistent spot unavailability
        if not has_spot:
            self.spot_unavailable_count += 1
        else:
            self.spot_unavailable_count = 0

        # Critical deadline: must use ON_DEMAND
        if remaining_task_time + self.restart_overhead >= remaining_time - 0.5:
            return ClusterType.ON_DEMAND

        slack_ratio = slack / max(remaining_task_time, 1e-6)

        if has_spot:
            if last_cluster_type == ClusterType.ON_DEMAND:
                # Break-even analysis: is switching to SPOT worth it?
                switch_cost = self.restart_overhead * 1.0
                savings_per_hour = 0.7  # OD(1.0) - SPOT(0.3)
                break_even = switch_cost / savings_per_hour
                if remaining_task_time < break_even * 1.5:
                    return ClusterType.ON_DEMAND
                if slack < self.restart_overhead * 3:
                    return ClusterType.ON_DEMAND
            return ClusterType.SPOT
        else:
            if last_cluster_type == ClusterType.ON_DEMAND:
                return ClusterType.ON_DEMAND
            # Graduated thresholds based on slack ratio
            if slack_ratio < 0.1:
                return ClusterType.ON_DEMAND
            if slack_ratio < 0.25 and self.spot_unavailable_count > 10:
                return ClusterType.ON_DEMAND
            if slack_ratio < 0.4 and self.spot_unavailable_count > 20:
                return ClusterType.ON_DEMAND
            return ClusterType.NONE  # Wait for spot
    # ... (remaining 15 lines: reset(), _from_args(),
    #  and additional small-task / tight-deadline guards)
\end{lstlisting}
\end{promptbox}

\subsection{CUDA Kernel: LayerNorm}
\label{app:sol-cuda}

We show the best individual kernel discovered for LayerNorm, which achieves a 3.32$\times$ speedup over the PyTorch baseline. The kernel employs three key techniques absent from the naive implementation: (1) \texttt{float4} vectorization that loads four values per memory transaction, cutting memory overhead by $\sim$4$\times$; (2) a two-pass algorithm (compute statistics, then normalize) that lets the GPU optimize each phase independently; and (3) warp shuffle reductions (\texttt{\_\_shfl\_down\_sync}) for direct register-to-register partial sum accumulation, bypassing slower shared memory paths. This kernel was discovered in multi-task mode, where optimization patterns transfer across the 31 KernelBench problems via the shared Pareto frontier.

\begin{promptbox}[Optimized LayerNorm CUDA Kernel (excerpt)]
\begin{lstlisting}[aboveskip=0pt,belowskip=0pt,numbers=none,frame=none,backgroundcolor=\color{gray!5},language=C++]
__inline__ __device__ float warp_sum(float v) {
    unsigned mask = 0xffffffffu;
    for (int offset = KB_WARP_SIZE / 2; offset > 0; offset >>= 1)
        v += __shfl_down_sync(mask, v, offset);
    return v;
}

__global__ void rowwise_stats_kernel(
    const float* __restrict__ x, float* __restrict__ mean,
    float* __restrict__ inv_std, int64_t B, int64_t M, float eps) {
    int64_t row = blockIdx.x;
    if (row >= B) return;
    const float* row_ptr = x + row * M;

    float thread_sum = 0.0f, thread_sumsq = 0.0f;
    // float4 vectorized loads when aligned
    const float4* row_v4 = reinterpret_cast<const float4*>(row_ptr);
    for (int64_t j = threadIdx.x; j < (M >> 2); j += blockDim.x) {
        float4 v = row_v4[j];
        thread_sum   += (v.x + v.y + v.z + v.w);
        thread_sumsq += (v.x*v.x + v.y*v.y + v.z*v.z + v.w*v.w);
    }
    // Warp shuffle reduction + cross-warp shared memory reduce
    thread_sum = warp_sum(thread_sum);
    thread_sumsq = warp_sum(thread_sumsq);
    // ... (shared memory cross-warp reduction, mean/inv_std output)
}

__global__ void layernorm_affine_kernel(
    const float* __restrict__ x, const float* __restrict__ weight,
    const float* __restrict__ bias, const float* __restrict__ mean,
    const float* __restrict__ inv_std, float* __restrict__ y,
    int64_t B, int64_t M) {
    int64_t row = blockIdx.x;
    float m = mean[row], inv = inv_std[row];
    // Vectorized normalize + affine transform
    const float4* x_v4 = reinterpret_cast<const float4*>(x + row*M);
    float4* y_v4 = reinterpret_cast<float4*>(y + row*M);
    for (int64_t j = threadIdx.x; j < (M >> 2); j += blockDim.x) {
        float4 xv = x_v4[j], wv = w_v4[j], bv = b_v4[j];
        y_v4[j] = {((xv.x-m)*inv)*wv.x+bv.x, ((xv.y-m)*inv)*wv.y+bv.y,
                    ((xv.z-m)*inv)*wv.z+bv.z, ((xv.w-m)*inv)*wv.w+bv.w};
    }
}
// ... (remaining 80 lines: host function, input validation,
//  thread/block configuration, Python module wrapper)
\end{lstlisting}
\end{promptbox}

\subsection{Circle Packing Algorithm}
\label{app:sol-circle}

The evolved circle packing algorithm (480+ lines) is a bilevel optimizer that jointly optimizes circle centers and radii for $n{=}26$ circles in a unit square. Starting from a simple greedy packing seed, the system discovers a multi-stage architecture: (1) an LP over radii with dual-variable sensitivities that provide exact gradients for center optimization, (2) L-BFGS-B over centers using these LP-derived gradients, (3) block SLP trust-region boosts targeting the worst-performing circles, (4) CMA-ES global exploration with automatic restarts, and (5) aggressive relocation of smallest circles to edges and corners. The algorithm also employs six diverse seeding strategies (hexagonal, uniform, edge-ring, farthest-point, corner-spokes, and edge-biased hex) to avoid local optima. We show the main entry point and key optimization components.

\begin{promptbox}[Evolved Circle Packing Algorithm (excerpt)]
\begin{lstlisting}[aboveskip=0pt,belowskip=0pt,numbers=none,frame=none,backgroundcolor=\color{gray!5}]
def main(timeout, current_best_solution):
    """Bilevel L-BFGS with exact LP sensitivities +
    SLP block boosts + CMA/Evolution fallback"""
    n = 26
    # LP for optimal radii given centers; returns duals
    def solve_radii_lp(centers, need_duals=False):
        # Boundary constraints: r_i <= min(x_i, 1-x_i, y_i, 1-y_i)
        # Pairwise constraints: r_i + r_j <= ||c_i - c_j||
        # Objective: maximize sum(r)
        res = linprog(c_obj, A_ub=A_ub, b_ub=b_ub, ...)
        return r, success, {'dual': res.ineqlin.marginals}

    # Gradient from LP duals: sensitivity of sum(r) to centers
    def gradient_from_duals(centers, dual_vec):
        # Boundary: g[i,0] += (dual_left - dual_right)
        # Pairwise: g[i] += lambda_k * unit_vec(i->j)
        return g

    # Bilevel: L-BFGS-B over centers, LP over radii
    def lbfgs_bilevel(centers_init, max_iters=300):
        def f_and_g(flat):
            r, _, info = solve_radii_lp(centers, need_duals=True)
            g = gradient_from_duals(centers, info['dual'])
            return -score, -g.reshape(-1)
        minimize(f_and_g, method='L-BFGS-B', bounds=bounds)

    # Block SLP: trust-region moves on worst circles
    def block_slp_boost(centers, rounds=4, k=10, delta=0.18):
        # Jointly optimize dx, dy, r via linearized LP
        # Focus on smallest radii + highest gradient circles

    # CMA-ES global exploration over center positions
    # 6 seeding strategies: hex, uniform, edge-ring,
    #   farthest-point, corner-spokes, edge-biased hex

    # Pipeline: seed evaluation -> L-BFGS bilevel ->
    #   block SLP boost -> CMA-ES -> relocate worst ->
    #   evolutionary exploration -> final SLP polish
    # ... (remaining 350 lines: seeding, CMA-ES, relocation,
    #  evolutionary exploration, final polish)
\end{lstlisting}
\end{promptbox}

\begin{figure*}[p]
    \centering

    \textbf{Zero-shot}\hspace{0.22\textwidth}\textbf{\oa{}}

    \vspace{0.4em}

    \includegraphics[width=0.495\textwidth,height=0.21\textheight,keepaspectratio]{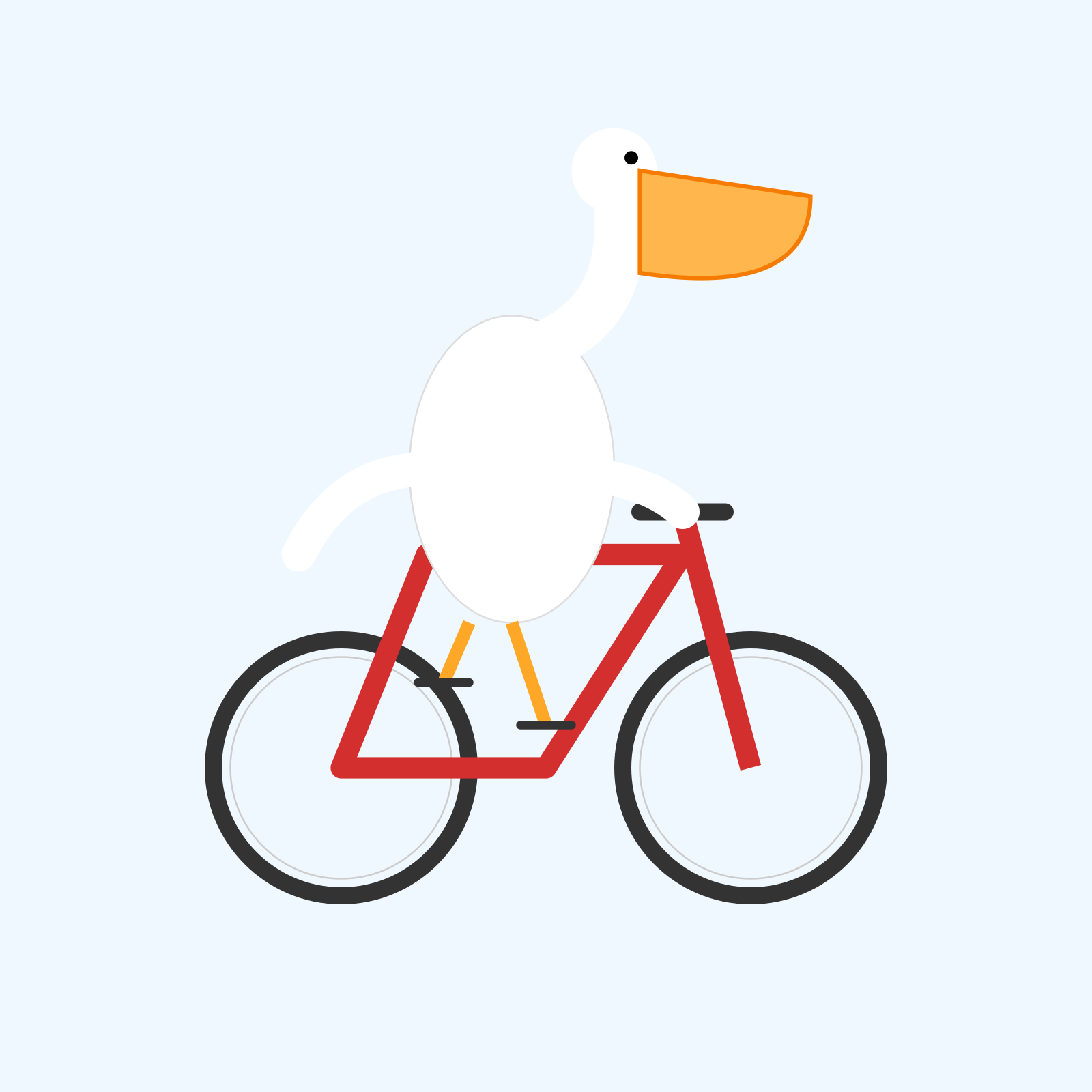}
    \hspace{0.01\textwidth}
    \includegraphics[width=0.495\textwidth,height=0.21\textheight,keepaspectratio]{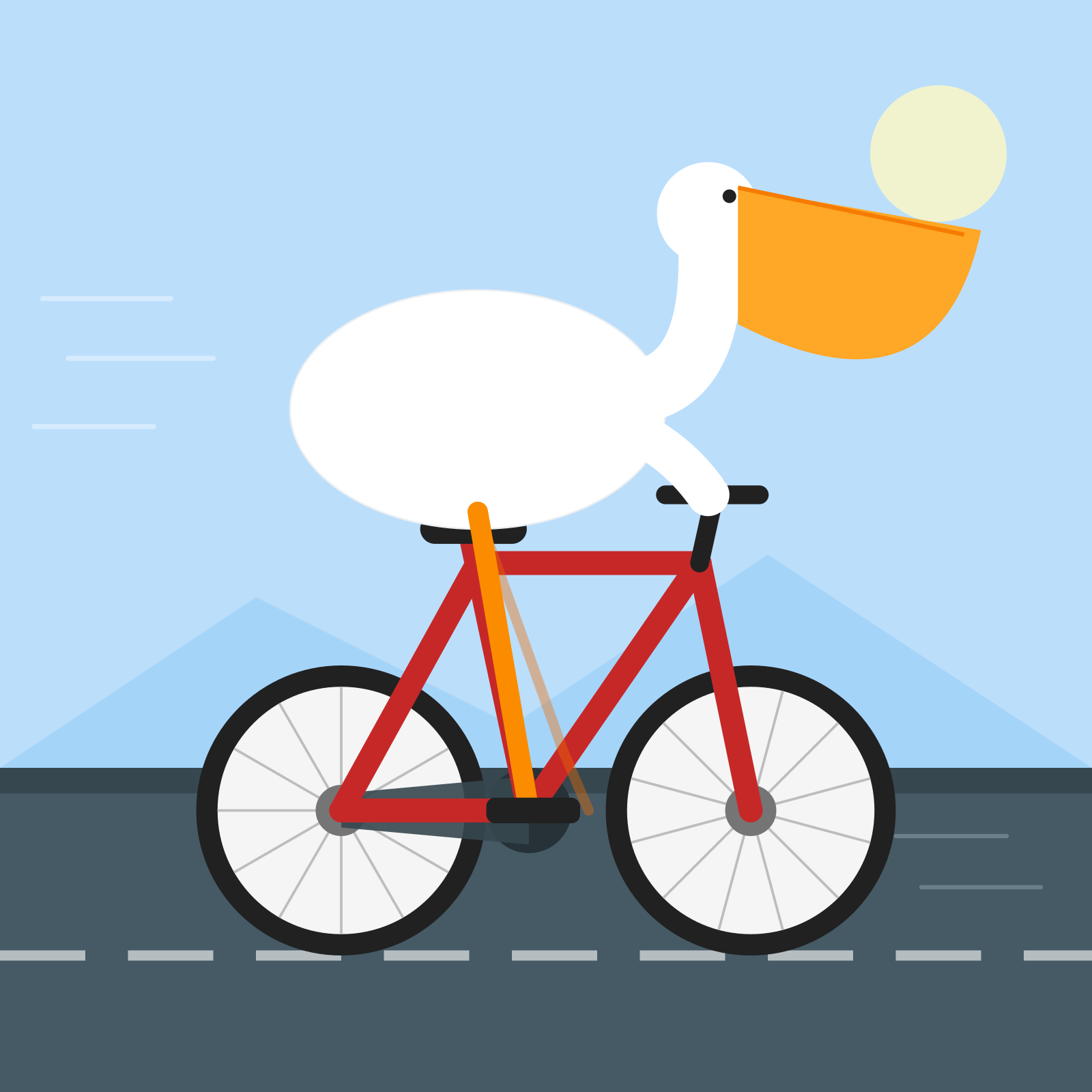}

    \vspace{0.35em}

    \includegraphics[width=0.495\textwidth,height=0.21\textheight,keepaspectratio]{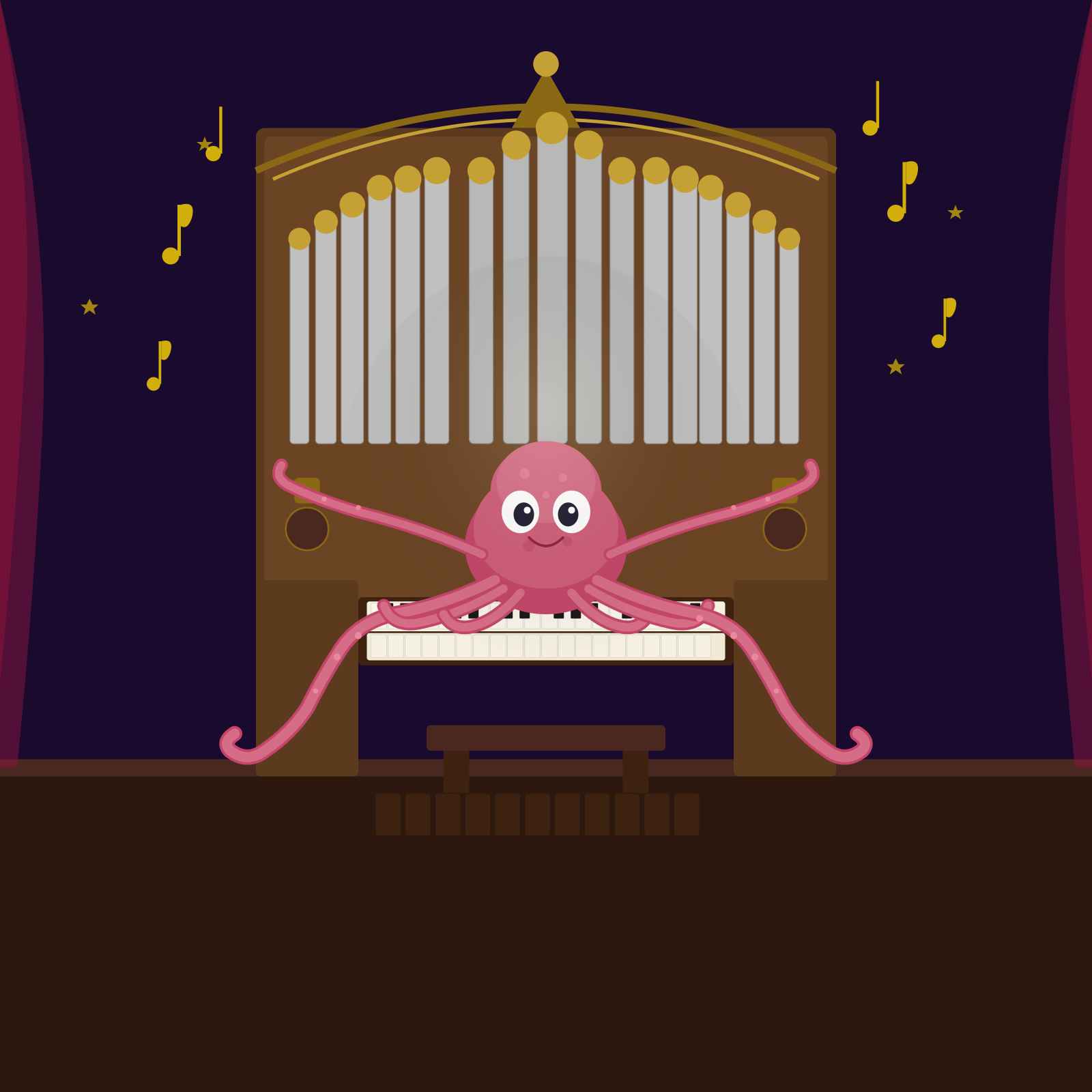}
    \hspace{0.01\textwidth}
    \includegraphics[width=0.495\textwidth,height=0.21\textheight,keepaspectratio]{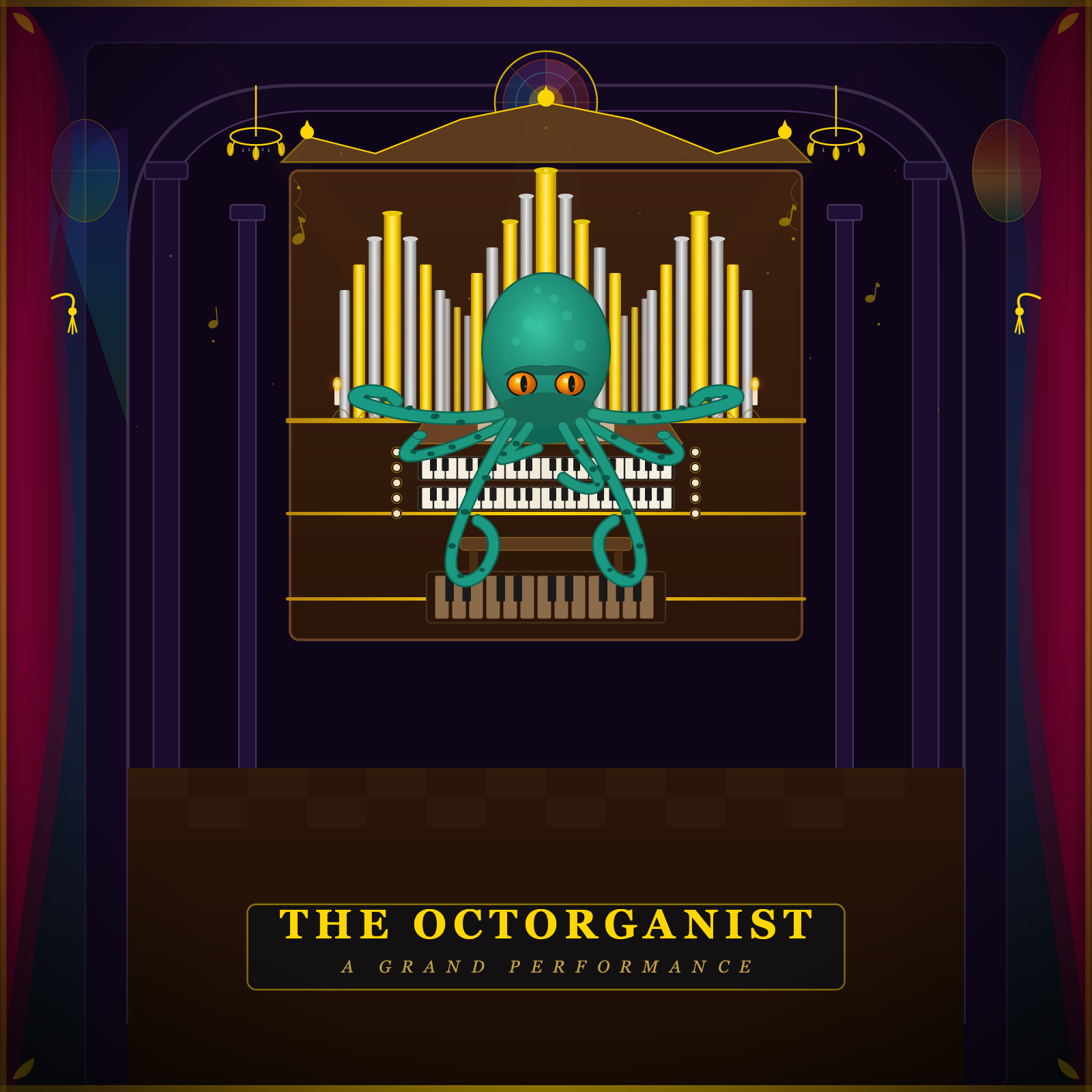}

    \vspace{0.35em}

    \includegraphics[width=0.495\textwidth,height=0.21\textheight,keepaspectratio]{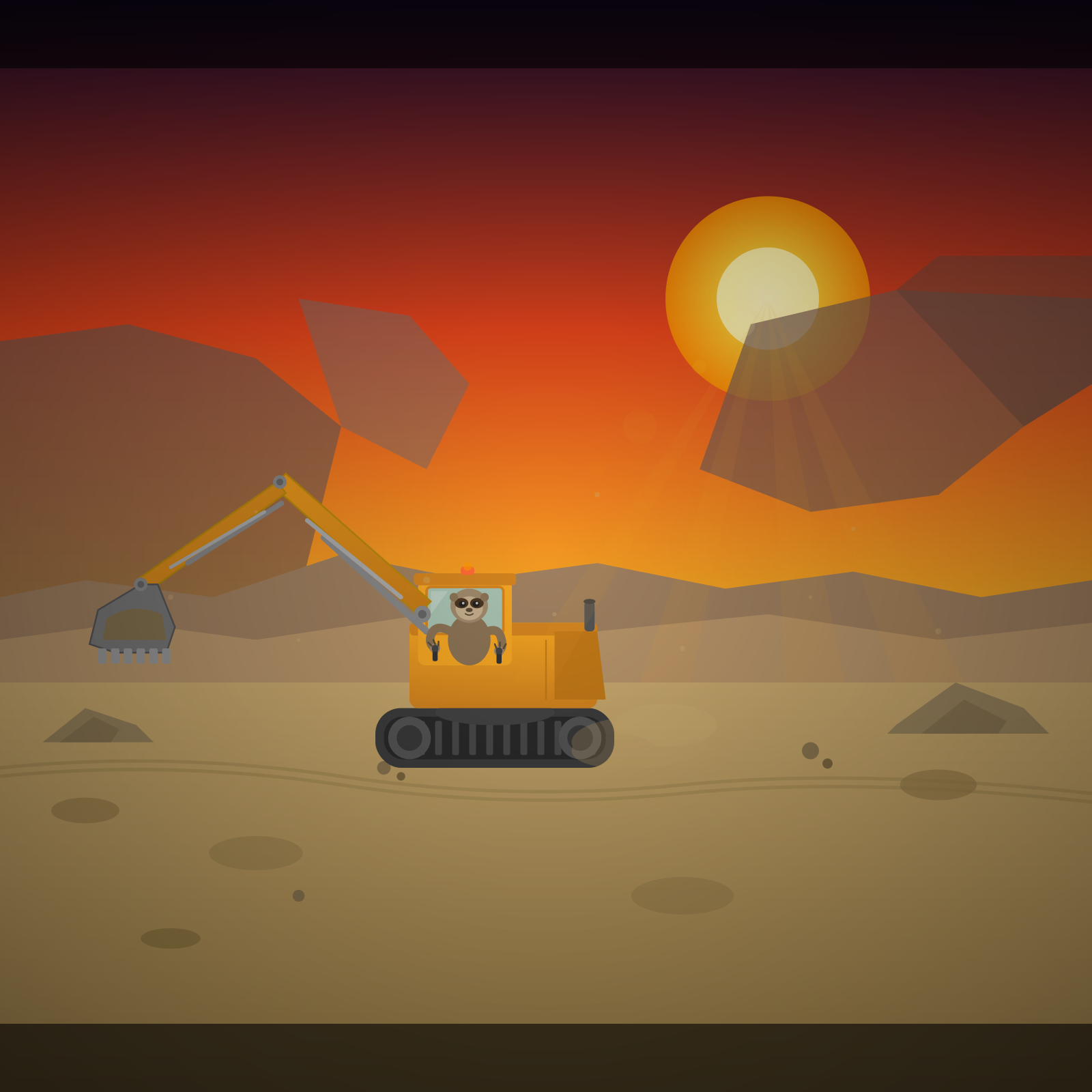}
    \hspace{0.01\textwidth}
    \includegraphics[width=0.495\textwidth,height=0.21\textheight,keepaspectratio]{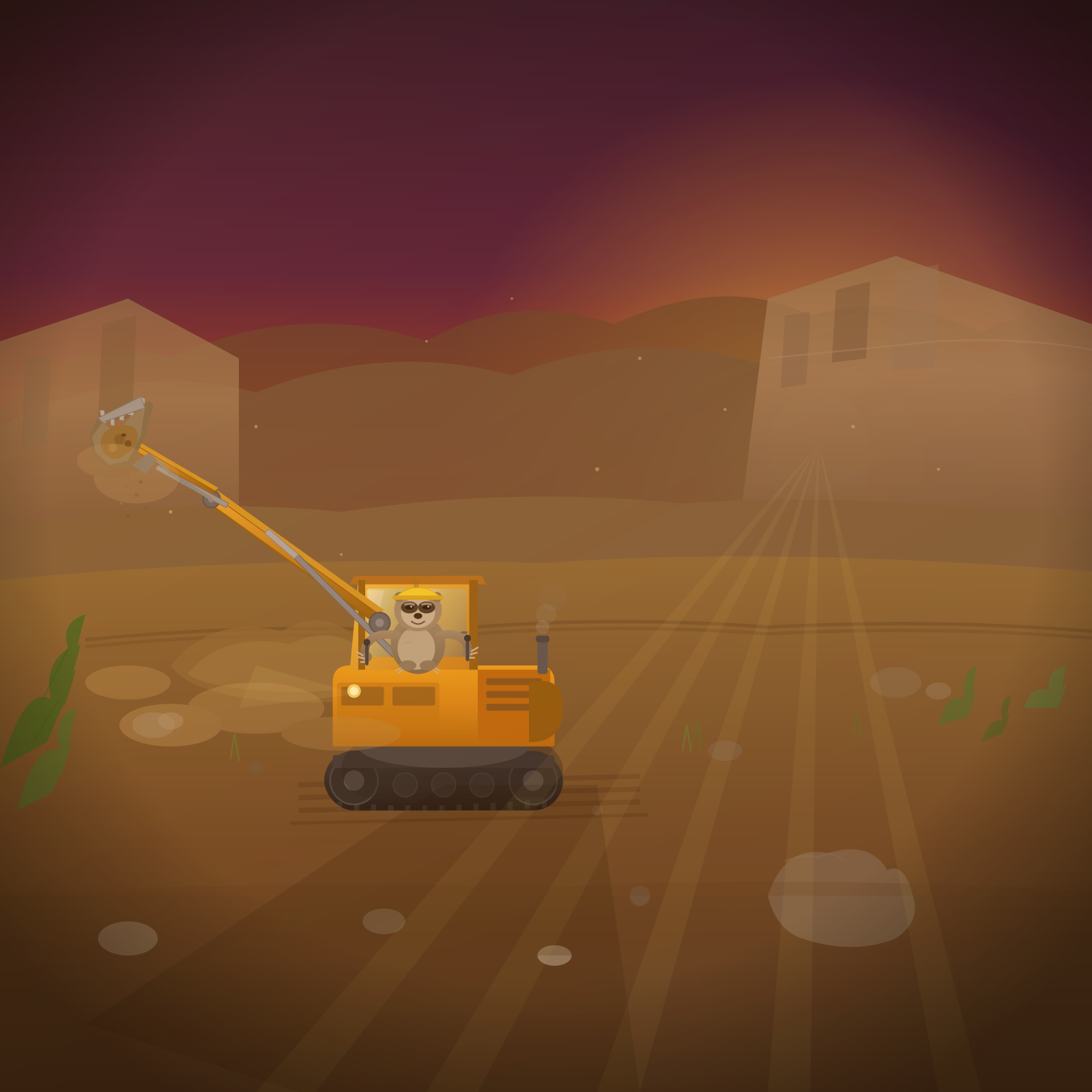}

    \vspace{0.35em}

    \includegraphics[width=0.495\textwidth,height=0.21\textheight,keepaspectratio]{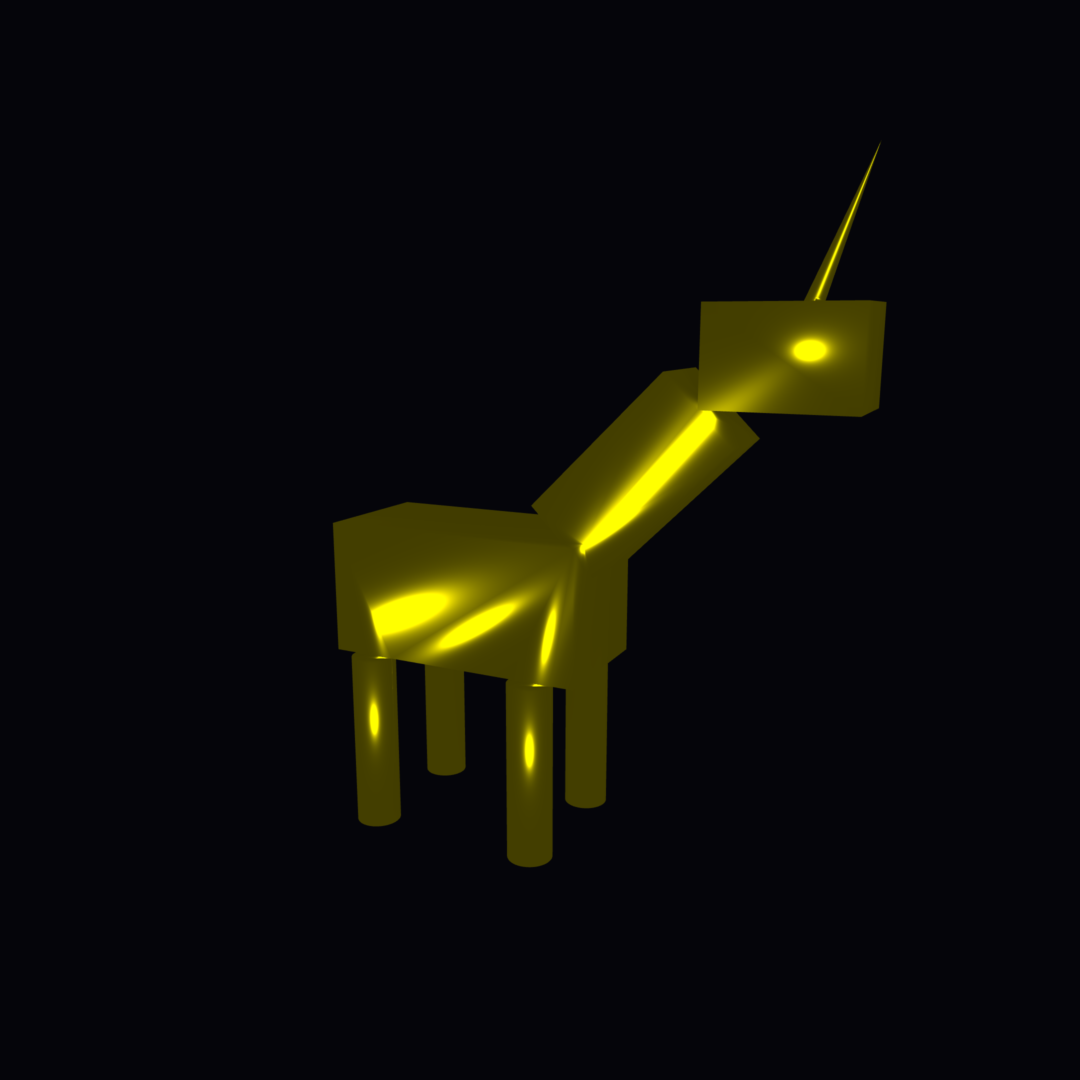}
    \hspace{0.01\textwidth}
    \includegraphics[width=0.495\textwidth,height=0.21\textheight,keepaspectratio]{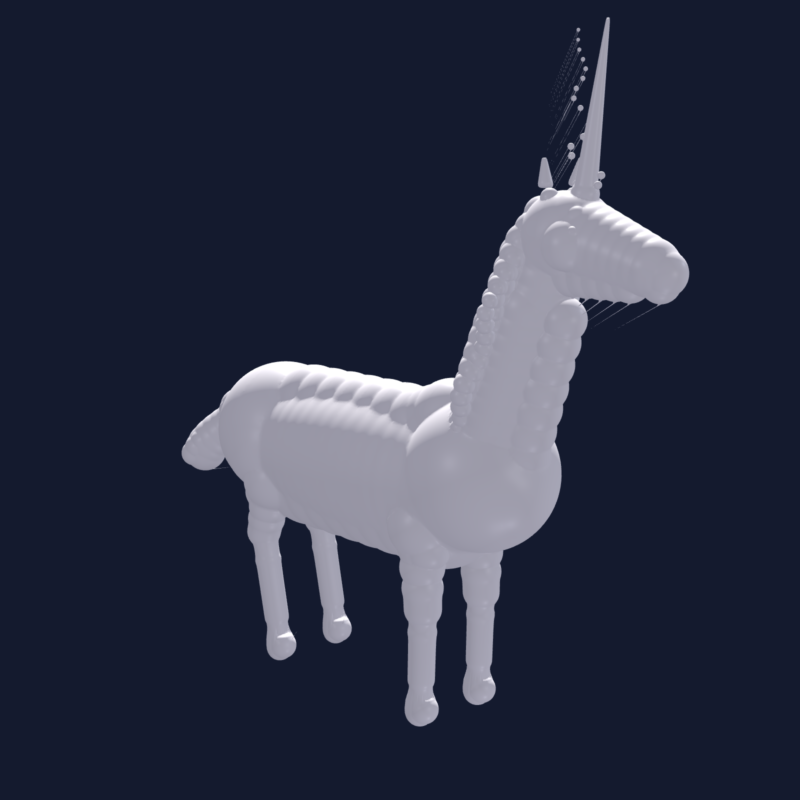}

    \caption{Qualitative comparison between zero-shot generations (left) and optimize\_anything candidates (right) across four example tasks. Optimization consistently improves many visual aspects including composition, structure, detail, and overall visual quality.}
    \label{fig:appendix_zero_shot_vs_oa}
\end{figure*}

\section{Demonstration}\label{app:demo}

A 4-minute demo video and accompanying artifacts are available at \url{https://drive.google.com/drive/folders/1mfd8xny_YRri5UYwTxKoBs3CJ_cpxpMr}. The demo showcases \oa{}'s generality through two end-to-end scenarios: evolving ARC-AGI agents and optimizing circle packing algorithms.

\paragraph{Scenario 1: Evolving ARC-AGI agents.} Starting from a naive 10-line agent (a single LLM call), \oa{} iteratively designs it into a 300+ line multi-stage pipeline with sub-agents, code generation, iterative debugging, and structured fallback logic. \asi{}---per-puzzle execution traces, error tracebacks, and model outputs---drives targeted architectural improvements. The final agent reaches 89.5\% accuracy on ARC-AGI~\cite{chollet2019arc} test puzzles using Gemini 3 Flash as both proposer and agent model (\S\ref{sec:arc}).

\paragraph{Scenario 2: Optimizing circle packing.} We demonstrate single-task search on packing $n{=}26$ circles in a unit square to maximize the sum of radii. \oa{} evolves a simple greedy packing seed into a 480+ line bilevel optimizer using LP-derived gradients and CMA-ES exploration, outperforming AlphaEvolve's~\cite{alphaevolve2025} reported solution (\S\ref{sec:circle}). The demo visualizes how the system discovers novel algorithmic components not present in the seed.

\paragraph{Live demonstration.} The demo runs both scenarios through Jupyter notebooks, allowing observation of optimization trajectories, inspection of intermediate candidates, and exploration of how diagnostic feedback drives improvements.

\section{Artifact Availability}\label{app:artifact}

\oa{} is open-sourced as part of the GEPA project. The source code is available at \url{https://github.com/gepa-ai/gepa}. A tutorial-style introduction is available at the accompanying blog post~\cite{agrawal2026oa_blog}. The complete reproduction artifact accompanying this paper is publicly available at \url{https://github.com/gepa-ai/optimize-anything-artifact} under the \texttt{acm\_cais\_artifact\_evaluation/} directory. Each evaluation
domain has its own subdirectory under \texttt{domains/} with runnable
\texttt{optimize\_anything} code, a \texttt{README.md} mapping the folder to the relevant section of this paper, and the saved \texttt{GEPAState} checkpoint from the paper run. See the top-level \texttt{README.md} for the reproduction guide.

\paragraph{Hardware notes.}
Most domains run on a single CPU host with API access to the proposer and
refiner LLMs (the paper used GPT-5/5.1, Gemini~3 Flash, and Claude Opus~4.6
depending on domain; exact identifiers are documented per domain). The
KernelBench domain requires an NVIDIA V100 32GB GPU with CUDA~12.1+.

\end{document}